\pdfoutput=1

\RequirePackage{fix-cm}

\documentclass[twocolumn]{svjour3}

\smartqed

\usepackage{epsfig}
\usepackage{graphicx}
\usepackage{amsmath}
\usepackage{amssymb}
\usepackage{mathtools}
\usepackage{makecell}
\usepackage{tabularx, booktabs}
\usepackage{cuted}
\usepackage{xcolor}
\usepackage{multirow}

\DeclareMathOperator*{\argmin}{argmin}
\newcommand*{\eqsep}{10pt}

\begin{document}

\title{SliderGAN: Synthesizing Expressive Face Images by Sliding 3D Blendshape Parameters.}

\author{
Evangelos Ververas $^\star$ \and
Stefanos Zafeiriou $^\dagger$}

\institute{    
$^\star$e.ververas16@imperial.ac.uk\\
$^\dagger$s.zafeiriou@imperial.ac.uk\\
\at
$^{\star,\dagger}$Queen’s Gate, London SW7 2AZ, UK\\
\at
$^{\dagger}$Center for Machine Vision and Signal Analysis, University of
Oulu, Oulu, Finland\\
}

\date{Received: date / Accepted: date}

\maketitle

\begin{abstract}

Image-to-image (i2i) translation is the dense regression problem of learning how to transform an input image into an output using aligned image pairs. Remarkable progress has been made in i2i translation with the advent of Deep Convolutional Neural Networks (DCNNs) and particular using the learning paradigm of Generative Adversarial Networks (GANs). In the absence of paired images, i2i translation is tackled with one or multiple domain transformations (i.e., CycleGAN, StarGAN etc.). In this paper, we study a new problem, that of image-to-image translation, under a set of continuous parameters that correspond to a model describing a physical process. In particular, we propose the SliderGAN which transforms an input face image into a new one according to the continuous values of a statistical blendshape model of facial motion. We show that it is possible to edit a facial image according to expression and speech blendshapes, using sliders that control the continuous values of the blendshape model. This provides much more flexibility in various tasks, including but not limited to face editing, expression transfer and face neutralisation, comparing to models based on discrete expressions or action units. 



   
\end{abstract}


\begin{figure*}
    \includegraphics[width=\textwidth]{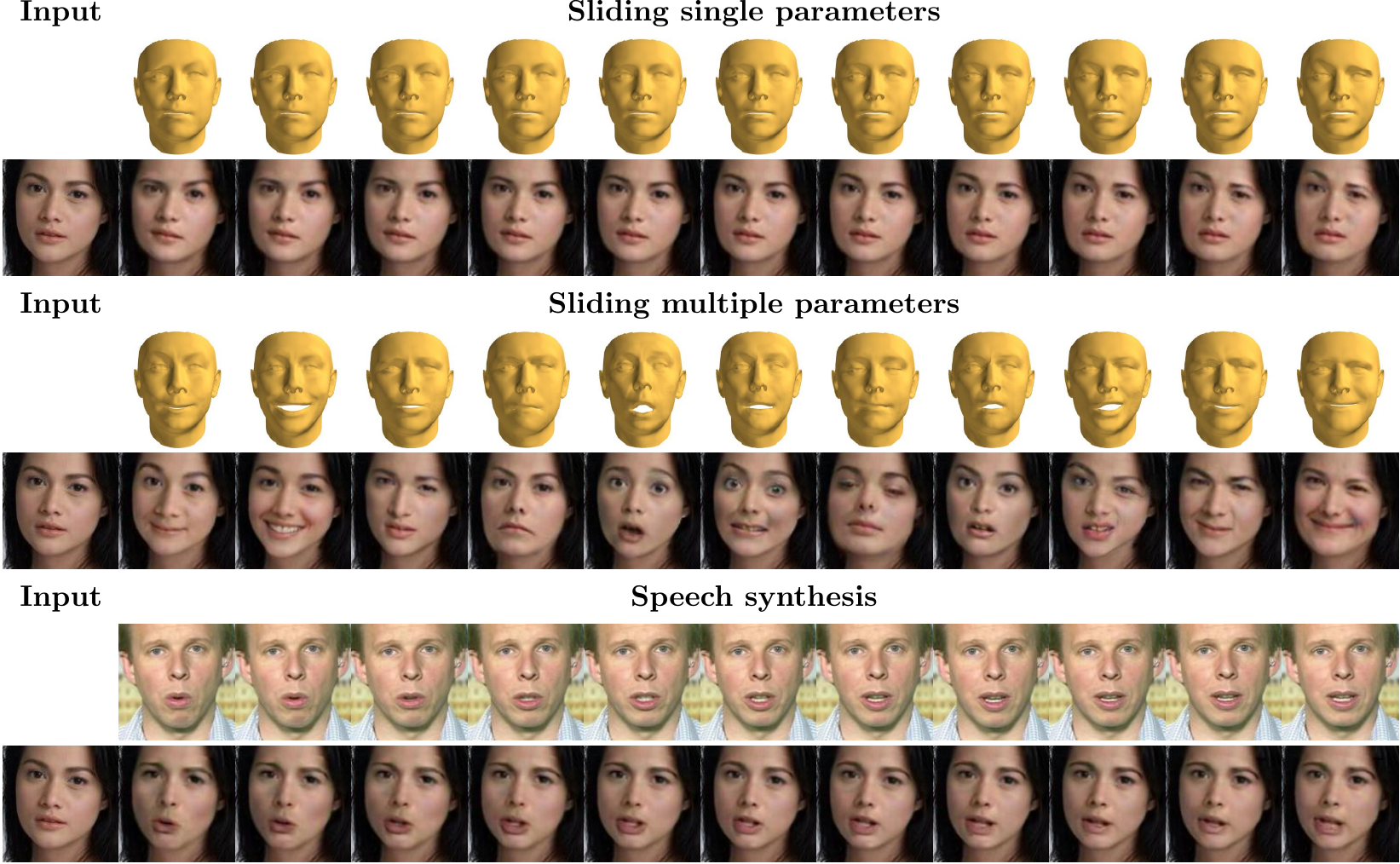}
    \caption{Expressive faces generated by sliding a single or multiple blendshape parameters in the normalized range $[-1, 1]$. Rows 1 and 3 depict 3D expressive faces generated by a linear blendshape model of natural face motion and a set of expression parameters. The corresponding edited images generated by SliderGAN using the same set of parameters are depicted in rows 2 and 4. As it is observed, the generated images accurately replicate the 3D faces' motion. The robustness of blendshape coding of facial motion allows SliderGAN to perform speech synthesis, as demonstrated in rows 5 (target speech) and 6 (synthesized speech), for which a 3D blenshape model of human speech was utilized.}
    \label{fig_single_param}
\end{figure*}

\section{Introduction}\label{section_intro}

Interactive editing of the expression of a face in an image has countless applications including but not limited to movies post-production, computational photography, face recognition (i.e. expression neutralisation) etc. In computer graphics facial motion editing is a popular field, nevertheless mainly revolves around constructing person-specific models having a lot of training samples \cite{suwajanakorn2017synthesizing}. Recently, the advent of machine learning, and especially Deep Convolutional Neural Networks (DCNNs) provide very exciting tools making the community to re-think the problem. In particular, recent advances in Generative Adversarial Networks (GANs) provide very exciting solutions for image-to-image (i2i) translation. 

i2i translation, i.e. the problem of learning how to transform aligned image pairs, has attracted a lot of attention during the last few years \cite{pix2pix2017,CycleGAN2017,StarGAN2018}. The so-called pix2pix model and alternatives demonstrated excellent results in image completion etc. \cite{pix2pix2017}. In order to perform i2i translation in absence of image pairs the so-called CycleGAN was proposed, which introduced a cycle-consistency loss \cite{CycleGAN2017}. CycleGAN could perform i2i translation between two domains only (i.e. in the presence of two discrete labels). The more recent StarGAN \cite{StarGAN2018} extended this idea further to accommodate multiple domains (i.e. multiple discrete labels). 

StarGAN can be used to transfer an expression to a given facial image by providing the discrete label of the target expression. Hence, it has quite small capabilities in expression editing and arbitrary expression transfer. The past year quite some deep learning related methodologies have been proposed for transforming facial images \cite{StarGAN2018,wiles2018x2face,pumarola2018ganimation}. The most closely related work to us is the recent work \cite{pumarola2018ganimation} that proposed the GANimation model. GANimation follows the same line of research as StarGAN to translate facial images according to the activation of certain facial Action Units (AUs)\footnote{AUs is a system to taxonomize motion of the human facial muscles \cite{ekman2002facial}.} and their intensities. Even though AU coding is a quite comprehensive model for describing facial motion, detecting AUs is currently an open problem both in controlled, as well as in unconstrained recording conditions\footnote{The state-of-the-art AU detection techniques achieve around 50\% F1 in EmotioNet challenge and from our experiments OpenFace \cite{amos2016openface} achieves lower than 20-25\%} \cite{benitez2018discriminant,benitez2017recognition}. In particular, in unconstrained conditions for certain AUs the detection accuracy is not high-enough yet \cite{benitez2018discriminant,benitez2017recognition}, which affects the generation accuracy of GANimation\footnote{The accuracy of the GANimation model is highly related to both the AU detection, as well as the estimation of their intensity, since the generator is jointly trained and influenced by a network that performs detection and intensity estimation.}. One of the reasons of the low accuracy of automatic annotation of AUs, is the lack of annotated data and the high cost of annotation which has to be performed by highly trained experts. Finally, even though AUs 10-28 model mouth and lip motion, only 10 of them can be automatically recognized (10, 12, 14, 15, 17, 20, 23, 25, 26, 28) which can only be achieved with low accuracy and thus, they cannot describe all possible lip motion patterns produced during speech. Hence, GANimation model cannot be used in straightforward manner for transferring speech.

In this paper, we are motivated by the recent successes in 3D face reconstruction methodologies from in-the-wild images \cite{richardson2017learning,tewari2017self,tran2018nonlinear,booth20183d,booth20173d}, which make use of a statistical model of 3D facial motion by means of a set of linear blenshapes, and propose a methodology for facial image translation using GANs driven by the continuous parameters of the linear blenshapes. The linear blendshapes can describe both the motion that is produced by expression \cite{ChengCVPR18} and/or motion that is produced by speech \cite{Tzirakis2019Synthesising3F}. On the contrary, neither discrete emotions nor facial action units can be used to describe the motion produced by speech or the combination of motion from speech and expression. We demonstrate that it is possible to transform a facial image along the continuous axis of individual expression and speech blendshapes. 

Moreover, contrary to StarGAN, which uses discrete labels regarding expression, and GANimation, which utilizes annotations with regards to action units, our methodology does not need any human annotations, as we operate using pseudo-annotations provided by fitting a 3D Morphable Model (3DMM) to images \cite{booth20183d} (for expression deformations) or by aligning audio signals \cite{Tzirakis2019Synthesising3F} (for speech deformations). Building on the automatic annotation process exploited by SliderGAN, a by-product of our training process is a very robust regression DCNN that estimates the blendshape parameters directly from images. This DCNN is extremely useful for expression and/or speech transfer as it can automatically estimate the blendshape parameters of target images.

i2i translation models have achieved photo-realistic results by utilizing different GAN optimization methods in literature. pix2pix employed the original GAN optimization technique proposed in \cite{NIPS2014_GANs}. However, the loss function of GAN may lead to the vanishing gradients problem during the learning process. Hence, more effective GAN frameworks emerged that were employed by i2i translation methods. CycleGAN uses LSGAN, which builds upon GAN adopting a least squares loss function for the discriminator. StarGAN and GANimation use WGAN-GP \cite{NIPS2017_WGANGP}, which enforces gradient clipping as a measure to regularize the discriminator. WGAN-GP, builds upon WGAN \cite{ArjovskyCB17} which minimizes an approximation of the Wasserstein distance to stabilize training of GANs.

A recent approach of efficient GAN optimization which has been used to produce higher quality textures \cite{wang2018esrgan}, is the Relativistic GAN (RGAN) \cite{jolicoeur-martineau2018}. RGAN was suggested in order to train the discriminator to simultaneously decrease the probability that real images are real, while increasing the probability that the generated images are real. In our work, we incorporate RGAN in the training process of SliderGAN and demonstrate that it can improve the generator which produces more detailed results in the task of i2i translation for expression and speech synthesis, when compared to training with WGAN-GP. In particular, we employ the Relativistic average GAN (RaGAN) which decides whether an image is relatively more realistic than the others on average, rather than whether it is real or fake. More details, as well as the benefits from this mechanism are presented in Section \ref{subsec_model}.


To summurize, the proposed method includes quite a few novelties. First of all, we showcase that SliderGAN is able to synthesize smooth deformations of expression and speech in images by utilizing 3D blendshape models of expression and speech respectively. Moreover, it is the first time to the best of our knowledge that a direct comparison of blendshape and AU coding is presented, for the task of expression and speech synthesis. In addition, our approach is annotation-free but offers much better accuracy that AUs-based methods. Furthermore, it is the first time that Relativistic GAN was employed for the task of expression and speech synthesis. We demonstrate in our results that SliderGAN trained with the RaGAN framework (SliderGAN-RaD) benefits towards producing more detailed textures, than when trained with the standard WGAN-GP framework (SliderGAN-WGP). Finally, we enhance the training of our model with synthesized data, leveraging the reconstruction capabilities of statistical shape models. 

    

\section{Face Deformation Modelling with Blendshapes}\label{section_blendshape_modelling}

\subsection{Expression Blendshape Models}\label{subsec_blendshapes}

\begin{figure}
    \includegraphics[width=0.48\textwidth]{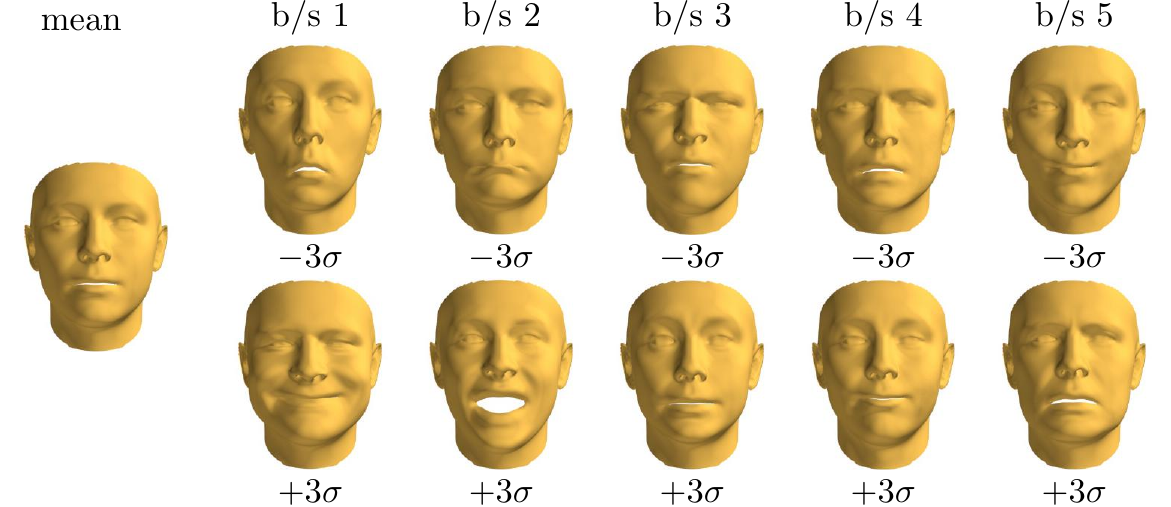}
    \caption{Visualization of the 5 most significant components of the blendshape model $\mathcal{S}_{exp}$. The 3D faces of this figure have been generated by adding the multiplied components to a mean face.}
    \label{fig_3D_bs_model}
\end{figure}

Blendshape models are frequently used in computer vision tasks as they constitute an effective parametric approach of modelling facial motion. The localized blendshape model \cite{neumann2013sparse} proposed a method to localize sparse deformation modes with intuitive visual interpretation. The model was built by sequences of manually collected expressive 3D face meshes. In more detail, a variant of sparse Principal Component Analysis (PCA) was applied to a matrix $\mathbf{D}=[\mathbf{d}_1, ..., \mathbf{d}_m] \in \mathbb{R}^{3n \times m}$, which includes $m$ difference vectors $\mathbf{d}_i \in \mathbb{R}^{3n}$, produced by subtracting each expressive mesh from the neutral mesh of each corresponding sequence. Therefore, the sparse blendshape components $\mathbf{C} \in \mathbb{R}^{h \times 1}$ where recovered by the following minimization problem:
\begin{equation}
\abovedisplayskip=\eqsep
\belowdisplayskip=\eqsep
    \argmin \|\mathbf{D} - \mathbf{B} \mathbf{C}\|_F^2 + \Omega(\mathbf{C}) \; \; \; \textup{s.t.} \; \mathcal{V}\left (  \mathbf{B} \right ), 
\label{eq:blendshape_spca}
\end{equation}
where, the constraint $\mathcal{V}$ can either be $\max \left ( \left | \mathbf{B}_{k} \right | \right ) = 1, \; \forall k$ or $\max \left ( \mathbf{B}_{k} \right ) = 1, \; \mathbf{B} \geq 1, \; \forall k$, with $\mathbf{B}_k \in \mathbb{R}^{3n \times 1}$ denoting the $k^{th}$ component of the sparse weight matrix $\mathbf{B} = [ \mathbf{B}_1, \cdots, \mathbf{B}_h]$. According to \cite{neumann2013sparse}, the selection of the constraints mainly controls whether face deformations will take place towards both negative and positive direction of the axes of the model's parameters or not, which is useful for describing shapes like muscle bulges. The regularization of sparse components $\mathbf{C}$ was performed with $\ell1 / \ell2$ norm~\cite{Wright2009,Bach2012}, while to compute optimal $\mathbf{C}$ and $\mathbf{B}$, an iterative alternating optimization was employed. The exact same approach was employed by \cite{ChengCVPR18}, in the construction of the 4DFAB blendshape model exploited in this work. The 5 most significant deformation components of the 4DFAB expression model are depicted in Fig. \ref{fig_3D_bs_model}.

\begin{figure}
    \includegraphics[width=0.48\textwidth]{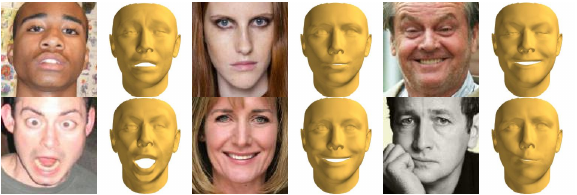}
    \caption{Examples of the 3D representation of the expression of an image by the model $\mathcal{S}_{exp}$. The 3D faces of this figure have been generated by 3DMM fitting on the corresponding images.}
    \label{fig_example_3Dexp_recon}
\end{figure}

\subsection{Extraction of expression parameters by 3DMM fitting}\label{subsec_3DMM_fitting}

3DMM fitting for 3D reconstruction of faces consists of optimizing three parametric models, the \textit{shape}, \textit{texture} and \textit{camera} models, in order to render a 2D instance as close as possible to the input image. To extract the expression parameters from an image we employ 3DMM fitting and particularly the approach proposed in \cite{booth20183d}.

In our pipeline we employ the identity variation of LSFM \cite{both2016lsfm}, which was learned from 10,000 face scans of unique identity, as the shape model to be optimized. To incorporate expression variation in the shape model, we combine LSFM with the 4DFAB blenshape model \cite{ChengCVPR18}, which was learned from 10,000 face scans of spontaneous and posed expression. The complete shape model can then be expressed as:
\begin{equation}
\abovedisplayskip=\eqsep
\belowdisplayskip=\eqsep
\begin{aligned}
    \mathcal{S}(\mathbf{p}_{id}, \mathbf{p}_{exp}) =& \ \mathbf{\bar{s}} + \mathbf{U}_{s,id}\mathbf{p}_{id} + \mathbf{U}_{s,exp}\mathbf{p}_{exp} 
    \\
    =& \ \mathbf{\bar{s}} + [\mathbf{U}_{s,id}, \mathbf{U}_{s,exp}][\mathbf{p}_{id}^\top, \mathbf{p}_{exp}^\top]^\top,
\end{aligned}
\label{eq_shape_model_detailed}
\end{equation}
where $\mathbf{\bar{s}}$ is the mean component of 3D shape, $\mathbf{U}_{s,id}$ and $\mathbf{U}_{s,expr}$ are the identity and expression subspaces of LSFM and 4DFAB respectively, and $\mathbf{p}_{id}$ and $\mathbf{p}_{expr}$ are the identity and expression parameters which are used to determine 3D shape instances.

Therefore, by fitting the 3DMM of \cite{booth20183d} in an input image $\mathbf{I}$, we can extract identity and expression parameters $\mathbf{p}_{id}$ and $\mathbf{p}_{exp}$ that instantiate the recovered 3D face mesh $\mathcal{S}(\mathbf{p}_{id}, \mathbf{p}_{exp})$. Based on the independent shape parameters for identity and expression, we exploit parameters $\mathbf{p}_{exp}$ to compose an annotated dataset of images and their corresponding vector of expression parameters $\{\mathbf{I}^{i}, \mathbf{p}_{exp}^{i}\}_{i=1}^{K}$, with no manual annotation cost.



\section{Proposed Methodology}\label{section_methodology}

In this section we develop the proposed methodology for continuous facial expression editing based on sliding the parameters of a 3D blendshape model.

\subsection{Slider-based Generative Adversarial Network for continuous facial expression and speech editing}\label{subsec_model}

\hspace{12pt}\textbf{Problem Definition} Let us here first formulate the problem under analysis and then describe our proposed approach to address it. We define an input image $\mathbf{I}_{org} \in \mathbb{R}^{H\times W\times 3}$ which depicts a human face of arbitrary expression. We further assume that any facial deformation or grimace evident in image $\mathbf{I}_{org}$, can be encoded by a parameter vector $\mathbf{p}_{org} = [p_{org,1}, p_{org,2}, ..., p_{org,N}]^\top$, of $N$ continuous scalar values $p_{org,i}$, normalized in the range $[-1, 1]$. In addition, the same vector $\mathbf{p}_{org}$ constitutes the parameters of a linear 3D blendshape model $\mathbf{S}_{exp}$ that, as in Fig. \ref{fig_example_3Dexp_recon}, instantiate the 3D representation of the facial deformation of image $\mathbf{I}_{org}$ which is  given by the expression:
\begin{equation}
\abovedisplayskip=\eqsep
\belowdisplayskip=\eqsep
    \mathcal{S}_{exp}(\mathbf{p}_{org}) = \mathbf{\bar{s}} + \mathbf{U}_{exp}\mathbf{p}_{org},
    \label{eq_3D_shape_instance}
\end{equation}
where $\mathbf{\bar{s}}$ is a mean 3D face component and $\mathbf{U}_{exp}$ the expression eigenbasis of the 3D blendshape model.

Our goal is to develop a generative model which given an input image $\mathbf{I}_{org}$ and a target expression parameter vector $\mathbf{p}_{trg}$, will be able to generate a new version $\mathbf{I}_{gen}$ of the input image with simulated expression given by the 3D expression instance $\mathcal{S}_{exp}(\mathbf{p}_{trg})$.


\smallskip
\textbf{Attention-Based Generator} To address the challenging problem described above, we propose to employ a Generative Adversarial Network architecture in order to train a generator network $\mathcal{G}$ that performs translation of an input image $\mathbf{I}_{org}$, conditioned on a vector of 3D blendshape parameters $\mathbf{p}_{trg}$; thus, learning the generator mapping $\mathcal{G}(\mathbf{I}_{org}|\mathbf{p}_{trg}) \rightarrow \mathbf{I}_{gen}$. In addition, to better preserve the content and the colour of the original images we employ an attention mechanism at the output of the generator as in \cite{NIPS2018_7627,pumarola2018ganimation}. That is we employ a generator with two parallel output layers, one producing a smooth deformation mask $\mathcal{G}_{m}  \in \mathbb{R}^{H\times W}$ and the other a deformation image $\mathcal{G}_i \in \mathbb{R}^{H\times W\times 3}$. The values of $\mathcal{G}_{m}$ are restricted in the region $[0, 1]$ by enforcing a sigmoid activation. Then, $\mathcal{G}_{m}$ and $\mathcal{G}_i$ are combined with the original image $\mathbf{I}_{org}$ to produce the target expression $\mathbf{I}_{gen}$ as:
\begin{equation}
\abovedisplayskip=\eqsep
\belowdisplayskip=\eqsep
    \mathbf{I}_{gen} = \mathcal{G}_{m}\mathcal{G}_i + (1 - \mathcal{G}_{m})\mathbf{I}_{org}.
\label{eq_att_mechanism}
\end{equation}

\smallskip
\textbf{Relativistic Discriminator} We employ a discriminator network $\mathcal{D}$ that forces the generator $\mathcal{G}$ to produce realistic images of the desired deformation. Different from the standard discriminator in GANimation which estimates the probability of an image being real, we employ the Relativistic Discriminator \cite{jolicoeur-martineau2018} which estimates the probability of an image being relatively more realistic than a generated one. That is if $\mathcal{D}_{img} = \sigma(\mathcal{C}(\mathbf{I}_{org}))$ is the activation of the standard discriminator, then $\mathcal{D}_{RaD,img} = \sigma(\mathcal{C}(\mathbf{I}_{org}) - \mathcal{C}(\mathbf{I}_{gen}))$ is the activation of the Relativistic Discriminator. Particularly, we employ the Relativistic average Discriminator (RaD) which accounts for all the real and generated data in a mini-batch. Then, the activation of the RaD is:
\begin{equation}
\centering
\abovedisplayskip=\eqsep
\belowdisplayskip=\eqsep
    \begin{aligned}
        & \mathcal{D}_{RaD,img} = \\
        & \hspace{-5pt} \left\{
        \begin{aligned}
            & \sigma(\mathcal{C}(\mathbf{I}) - \mathbb{E}_{I_{gen}}[\mathcal{C}(\mathbf{I}_{gen})]), \mbox{if $\mathbf{I}$ is a real image}
            \\
            & \sigma(\mathcal{C}(\mathbf{I}) - \mathbb{E}_{I_{org}}[\mathcal{C}(\mathbf{I}_{org})]) , \mbox{if $\mathbf{I}$ is a generated image}
        \end{aligned}
        \right.
    \end{aligned}
\label{eq_RaD}
\end{equation}
where $\mathbb{E}_{I_{org}}$ and $\mathbb{E}_{I_{gen}}$ define the average activations of all real and generated images in a mini-batch respectively. 

We further extend $\mathcal{D}$ by adding a regression layer parallel to $\mathcal{D}_{img}$ that estimates a parameter vector $\mathbf{p}_{est}$, to encourage the generator to produce accurate facial expressions, $\mathcal{D}(\mathbf{I}) \rightarrow \mathcal{D}_p(\mathbf{I}) = \mathbf{p}_{est}$. Finally, we aim to boost the ability of $\mathcal{G}$ to maintain face identity between the original and the generated images by incorporating a face recognition module $\mathcal{F}$.

\begin{figure*}
    \centering
    \includegraphics[width=\textwidth]{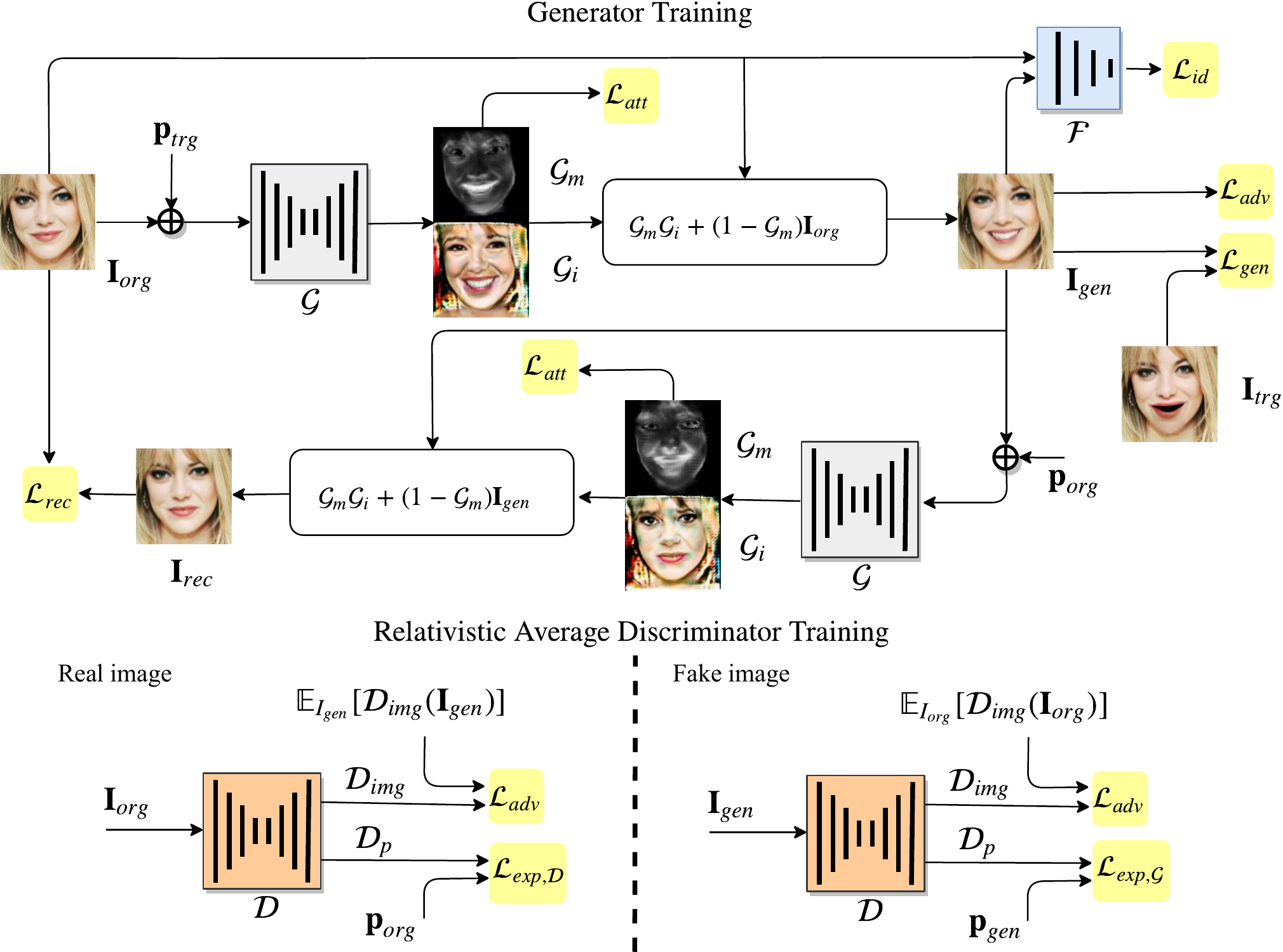}
    \caption{Synopsis of the modules, losses and the training process of SliderGAN. A attention-based generator $\mathcal{G}$ is trained to generate realistic expressive faces from continuous parameters by employing a set of adversarial, generation, reconstruction, identity and attention losses. The performance of our model is significantly boosted by employing synthetic image pairs through the $\mathcal{L}_{gen}$ loss. Moreover, a relativistic discriminator $\mathcal{D}$ is trained to classify images as relatively more real or fake, as well as to regress expression parameters of the input images in order to increase the generation quality of $\mathcal{G}$.}
    \label{fig_model}
\end{figure*}

\smallskip
\textbf{Semi-supervised training} We train our model in a semi-supervised manner with both data with no image pairs of the same person under different expressions $\{\mathbf{I}_{org}^{i}, \mathbf{p}_{org}^{i}, \mathbf{p}_{trg}^{i}\}_{i=1}^{K}$ and data with image pairs that we automatically generate as described in detail in Section \ref{subsec_data}, $\{\mathbf{I}_{org}^{i}, \mathbf{p}_{org}^{i}, \mathbf{I}_{trg}^{i}, \mathbf{p}_{trg}^{i}\}_{i=1}^{L}$. The modules of our model, as well as the training process of SliderGAN are presented in Fig. \ref{fig_model}.

\smallskip
\textbf{Adversarial Loss} To improve the photorealism of our synthesized images we utilize the Wasserstein GAN adversarial objective with gradient penalty (WGAN-GP) \cite{NIPS2017_WGANGP}.  Therefore, the selected WGAN-GP adversarial objective with RaD is defined as:
\begin{equation}
\centering
\abovedisplayskip=\eqsep
\belowdisplayskip=\eqsep
\begin{aligned}
    \mathcal{L}_{adv} = & \ \mathbb{E}_{I_{org}}[\mathcal{D}_{RaD,img}(\mathbf{I}_{org})] \\
    &- \mathbb{E}_{I_{org},p_{trg}}[\mathcal{D}_{RaD,img}(\mathcal{G}(\mathbf{I}_{org}, \mathbf{p}_{trg}))] \\
    &- \lambda_{gp}\mathbb{E}_{I_{gen}}[(\|\nabla_{I_{org}}\mathcal{D}_{img}(\mathbf{I}_{gen})\|_2 - 1)^2].
\end{aligned}
\label{eq_gan_loss}
\end{equation}
Different from the standard discriminator, both real and generated images are included in the generator part of the objective of Eq. \ref{eq_gan_loss}. This allows the generator to benefit by the gradients of both real and fake images, which as we show in experimental section leads to generated images with sharper edges and more details which also better represent the distribution of the real data.

Based on the original GAN rational \cite{NIPS2014_GANs} and the Relativistic GAN \cite{jolicoeur-martineau2018}, our generator $\mathcal{G}$ and discriminator $\mathcal{D}$ are involved in a min-max game, where $\mathcal{G}$ tries to maximize the objective of Eq.(\ref{eq_gan_loss}) by generating realistic images to fool the discriminator, while $\mathcal{D}$ tries to minimize it by correctly classifying real images as more realistic than fake and generated images as less realistic than real.

\smallskip
\textbf{Expression Loss} To make $\mathcal{G}$ consistent in accurately transferring target deformations $\mathcal{S}_{exp}(\mathbf{p}_{trg})$ to the generated images, we consider the discriminator $\mathcal{D}$ to have the role of an inspector. To this end, we back-propagate a mean squared loss between the estimated vector $\mathbf{p}_{est}$ of the regression layer of $\mathcal{D}$ and the actual vector of expression parameters of an image.

We apply the expression loss both for original images and generated ones. Similarly to the classification loss of StarGAN \cite{StarGAN2018}, we construct separate losses for the two cases. For real images $\mathbf{I}_{org}$ we define the loss:
\begin{equation}
\abovedisplayskip=\eqsep
\belowdisplayskip=\eqsep
    \mathcal{L}_{exp,\mathcal{D}} = \frac{1}{N}\|\mathcal{D}(\mathbf{I}_{org}) - \mathbf{p}_{org})\|^2,
\label{eq_exp_D_loss}
\end{equation}
between the estimated and real expression parameters of $\mathbf{I}_{org}$, while for the generated images we define the loss: 
\begin{equation}
\abovedisplayskip=\eqsep
\belowdisplayskip=\eqsep
    \mathcal{L}_{exp,\mathcal{G}} = \frac{1}{N}\|\mathcal{D}(\mathcal{G}(\mathbf{I}_{org}, \mathbf{p}_{trg})) - \mathbf{p}_{trg})\|^2,
\label{eq_exp_G_loss}
\end{equation}
between the estimated and target expression parameters of $\mathbf{I}_{gen} = \mathcal{G}(\mathbf{I}_{org}, \mathbf{p}_{trg})$. Consequently, $\mathcal{D}$ minimizes $\mathcal{L}_{exp,\mathcal{D}}$ to accurately regress the expression parameters of real images, while $\mathcal{G}$ minimizes $\mathcal{L}_{exp,\mathcal{G}}$ to generate images with accurate expression according to $\mathcal{D}$.

\smallskip
\textbf{Image Reconstruction Loss} The adversarial and the expression loss of Eq.(\ref{eq_gan_loss}) and Eq.(\ref{eq_exp_D_loss}), Eq.(\ref{eq_exp_G_loss}) respectively, would be enough to generate random realistic expressive images which however, would not preserve the contents of the input image $\mathbf{I}_{org}$. To overcome this limitation we admit a cycle consistency loss \cite{CycleGAN2017} for our generator $\mathcal{G}$:
\begin{equation}
\abovedisplayskip=\eqsep
\belowdisplayskip=\eqsep
    \mathcal{L}_{rec} = \frac{1}{W \times H}\|\mathbf{I}_{org} - \mathbf{I}_{rec}\|_1,
\label{eq_rec_loss}
\end{equation}
over the vectorized forms of the original image $\mathbf{I}_{org}$ and the reconstructed image $\mathbf{I}_{rec} = \mathcal{G}(\mathcal{G}(\mathbf{I}_{org}, \mathbf{p}_{trg}), \mathbf{p}_{org})$. Note that we obtain image $\mathbf{I}_{rec}$ by using the generator twice, first to generate image $\mathbf{I}_{gen} = \mathcal{G}(\mathbf{I}_{org}, \mathbf{p}_{trg})$ and then to get the reconstructed $\mathbf{I}_{rec} = \mathcal{G}(\mathbf{I}_{gen}, \mathbf{p}_{org})$, conditioning $\mathbf{I}_{gen}$ on the parameters $\mathbf{p}_{org}$ of the original image. 

\smallskip
\textbf{Image Generation Loss} To further boost our generator towards accurately transferring the expression from a vector of parameters to the edited image, we introduce image pairs of the form $\{\mathbf{I}_{org}^{i}, \mathbf{p}_{org}^{i}, \mathbf{I}_{trg}^{i}, \mathbf{p}_{trg}^{i}\}_{i=1}^{L}$ that we automatically generate from neutral images as described in detail in Section \ref{subsec_data}. We exploit the synthetic pairs of images of the same individualss under different expression by introducing an image generation loss:
\begin{equation}
\abovedisplayskip=\eqsep
\belowdisplayskip=\eqsep
    \mathcal{L}_{gen} = \frac{1}{W \times H}\|\mathbf{I}_{trg} - \mathbf{I}_{gen}\|_1,
\label{eq_gen_loss}
\end{equation}
where $\mathbf{I}_{trg}$ and $\mathbf{I}_{gen}$ are images with either neutral or synthetic expression of the same individual. Here, we calculate the $L1$ loss between the synthetic ground truth image $\mathbf{I}_{trg}$ and the generated by $\mathcal{G}$, $\mathbf{I}_{gen}$, aiming to boost our generator to accurately transfer the 3D expression $\mathcal{S}_{exp}(\mathbf{p}_{trg})$ to the edited image.

\smallskip
\textbf{Identity Loss} Image reconstruction loss of Eq.(\ref{eq_rec_loss}), aids to maintain the surroundings between the original and generated images. However, the faces' identity is not always maintained by this loss, as also show by our ablation study in Section \ref{subsec_ablation}. To alleviate this issue, we introduce a face recognition loss adopted from ArcFace \cite{deng2018arcface}, which models face recognition confidence by an angular distance loss. Particularly, we introduce the loss:
\begin{equation}
\abovedisplayskip=\eqsep
\belowdisplayskip=\eqsep
    \mathcal{L}_{id} = 1 - \cos(\mathbf{e}_{gen}, \mathbf{e}_{org}) = 1 - \frac{\|\mathbf{e}_{gen}\|\|\mathbf{e}_{org}\|}{\mathbf{e}_{gen}^\top\mathbf{e}_{org}},
\label{eq_id_loss}
\end{equation}
where $\mathbf{e}_{gen} = \mathcal{F}(\mathbf{I}_{gen})$ and $\mathbf{e}_{org} = \mathcal{F}(\mathbf{I}_{org})$ are embeddings of $\mathbf{I}_{gen}$ and $\mathbf{I}_{org}$ respectively, extracted by the face recognition module $\mathcal{F}$. According to ArcFace, face verification confidence is higher as the cosine distance $\cos(\mathbf{e}_{gen}, \mathbf{e}_{org})$ grows. During training, $\mathcal{G}$ is optimized to maintain face identity between $\mathbf{I}_{gen}$ and $\mathbf{I}_{org}$ which minimizes Eq.(\ref{eq_id_loss}). 

\smallskip
\textbf{Attention Mask Loss} To encourage the generator to produce sparse attention masks $\mathcal{G}_{m}$ that focus on the deformation regions and do not saturate to 1, we employ a sparsity loss $\mathcal{L}_{att}$. That is we calculate and minimize the $L1$-norm of the produced masks for both the generated and the reconstructed images, defining the loss as:
\begin{equation}
\abovedisplayskip=\eqsep
\belowdisplayskip=\eqsep
    \mathcal{L}_{att} = \frac{1}{W \times H}\Big(\|\mathcal{G}_{m}(\mathbf{I}_{org}, \mathbf{p}_{trg})\|_1 + \|\mathcal{G}_{m}(\mathbf{I}_{gen}, \mathbf{p}_{org})\|_1\Big),
\label{eq_att_loss}
\end{equation}

\smallskip
\textbf{Total Training Loss} We combine loss functions of Eq.(\ref{eq_gan_loss}) - Eq.(\ref{eq_att_loss}) to form loss functions $\mathcal{L}_{\mathcal{G}}$ and $\mathcal{L}_{\mathcal{D}}$ for separately training the generator $\mathcal{G}$ and the discriminator $\mathcal{D}$ of our model. We formulate the loss functions as:
\begin{equation}
    \begin{aligned}
    &\mathcal{L}_{\mathcal{G}} = \\
    &\left\{ 
    \begin{aligned}
        \mathcal{L}_{adv} + \lambda_{exp}\mathcal{L}_{exp,\mathcal{G}} + \lambda_{rec}\mathcal{L}_{rec} + \lambda_{id}\mathcal{L}_{id} + \lambda_{att}\mathcal{L}_{att}, &
        \\
        \mbox{for unpaired data } \{\mathbf{I}_{org}^{i}, \mathbf{p}_{org}^{i},  \mathbf{p}_{trg}^{i}\}_{i=1}^{K} &
        \\
        \mathcal{L}_{adv} + \lambda_{exp}\mathcal{L}_{exp,\mathcal{G}} + \lambda_{rec}\mathcal{L}_{rec} + \lambda_{gen}\mathcal{L}_{gen} + \lambda_{id}\mathcal{L}_{id}, & 
        \\
        + \lambda_{att}\mathcal{L}_{att}, \mbox{for paired data } \{\mathbf{I}_{org}^{i}, \mathbf{p}_{org}^{i}, \mathbf{I}_{trg}^{i}, \mathbf{p}_{trg}^{i}\}_{i=1}^{L} &
    \end{aligned}
    \right.
    \end{aligned}
\label{eq_G_loss}
\end{equation}
\begin{equation}
    \mathcal{L}_{\mathcal{D}} = - \mathcal{L}_{adv} + \lambda_{exp}\mathcal{L}_{exp,\mathcal{D}},
\label{eq_D_loss}
\end{equation}
where $\lambda_{exp}$, $\lambda_{rec}$, $\lambda_{gen}$, $\lambda_{id}$ and $\lambda_{att}$ are parameters that regularize the importance of each term in the total loss function. We discuss the choice of those parameters in Section \ref{subsec_training}.

As can be noticed in Eq.(\ref{eq_G_loss}), we employ different loss functions $\mathcal{L}_{\mathcal{G}}$, depending on if the training data are the real data with no image pairs or the synthetic data which include pairs. The only difference is that in the case of paired data we use the additional supervised loss term $\mathcal{L}_{gen}$.

\subsection{Implementation and training details}\label{subsec_training}

Having presented the architecture of our model, here we report further implementation and training details. For the generator module $\mathcal{G}$ of SliderGAN, we adopted the architecture of CycleGAN \cite{CycleGAN2017} as it is proved to generate remarkable results in image-to-iamge translation problems, as for example in StarGAN \cite{StarGAN2018}. We extended the generator by adding a parallel output layer to accomodate the attention mask mechanism. Moreover, for $\mathcal{D}$ we adopted the architecture of PatchGAN \cite{pix2pix2017} which produces probability distributions of the multiple image patches to be real or generated, $\mathcal{D}(\mathbf{I}) \rightarrow \mathcal{D}_{img}$. As described in Section \ref{subsec_model}, we extended this discriminator architecture by adding a parallel regression layer to estimate continuous expression parameters.

We trained our model with images of size $128 \times 128$, aligned to a reference shape of 2D landmarks. As condition vectors for our experiments, we utilized the 30 most significant expression components of 4DFAB and the 10 most significant speech components of LRW-3D \cite{Tzirakis2019Synthesising3F}. We set the batch size to 16 and trained our model for 60 epochs with Adam \cite{Adam} ($\beta_1 = 0.5, \beta_2 = 0.999$). We first trained our model only with the generated image pairs for 20 epochs and then proceeded to unsupervised training for another 40 epochs with unpaired images. Lastly, we chose loss weights $\lambda_{adv} = 30$, $\lambda_{exp} = 1000$, $\lambda_{rec} = 10$, $\lambda_{gen} = 10$, $\lambda_{id} = 4$ and $\lambda_{att} = 0.3$. Larger values for $\lambda_{id}$ significantly restrict $\mathcal{G}$, driving it to generate images very close to the original ones with no change in expression. Also, lower values for $\lambda_{att}$, lead to mask saturation.

\section{Experiments}\label{section_experiments}

In this section we present a series of experiments that we conducted in order to evaluate the performance of SliderGAN. First, we describe the datasets we utilized to train and test our model (Section \ref{subsec_data}). Then, we test the ability of SliderGAN to manipulate the expression in images by adjusting a single or multiple parameters of a 3D blendshape model (Section \ref{subsec_3dbased}). Moreover, we present our results in direct expression transfer between an input and a target image (Section \ref{subsec_exprtrsf}) and in discrete expression synthesis (Section \ref{subsec_six_basic}). We examine the ability of SliderGAN to handle face deformations due to speech (Section \ref{subsec_speech}) and test the regression accuracy of our model's discriminator (Section \ref{subsec_expr3Drec}). We close the experimental section of our work by presenting an ablation study on the contribution of the different loss functions of our technique (Section \ref{subsec_ablation}).

\subsection{Datasets}\label{subsec_data}

\begin{figure}
    \includegraphics[width=0.49\textwidth]{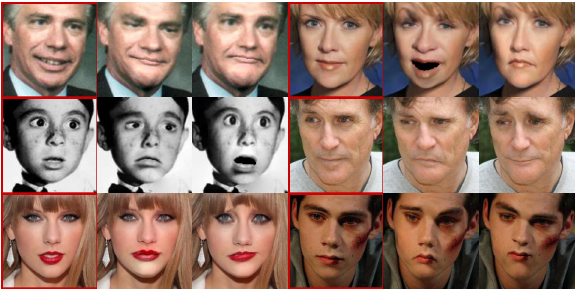}
    \caption{Synthetic expressive faces, generated by fitting a 3DMM on the original images and rendering back with a randomly sampled expression. The images with a red frame are the original images.}
    \label{fig_rendered_data}
\end{figure}

\begin{figure*}
    \includegraphics[width=\textwidth]{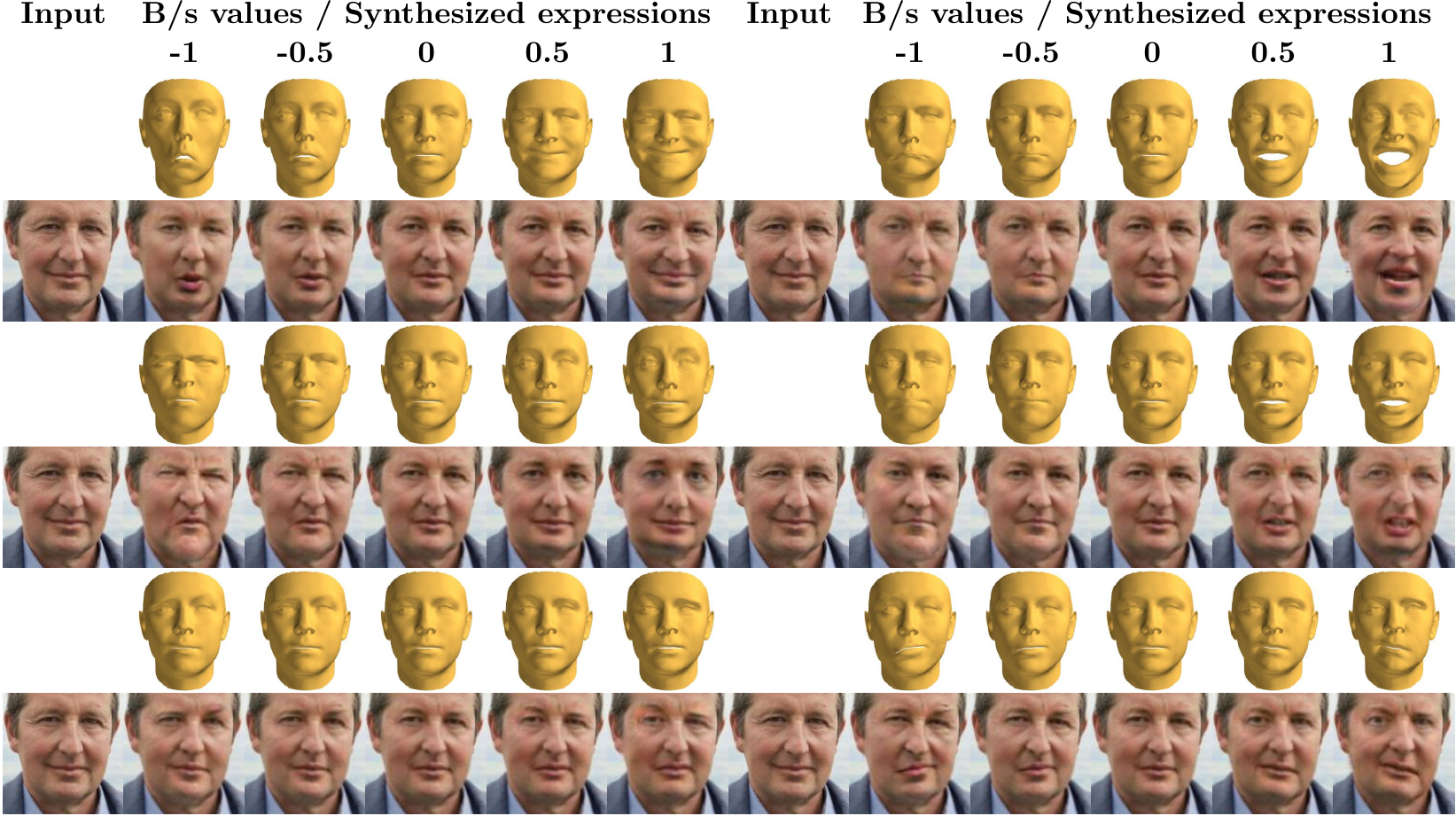}
    \caption{Expressive faces generated by sliding single blendshape (b/s) parameters in the range $[-1, 1]$. As it is observed, the edited images accurately replicate the 3D faces' motion in the whole range of parameter values.}
    \label{fig_single_param}
\end{figure*}

\hspace{12pt} \textbf{Emotionet} For the training and validation phases of our algorithm we utilized a subset of 250,000 images of the EmotioNet database \cite{EmotioNet2016}, which contains over 1 million images of expression and emotion, accompanied by annotations about facial Action Units. However, SliderGAN is trained with image - blenshape parameters pairs which are not available. Therefore, in order to extract the expression parameters we fit the 3DMM of \cite{booth20183d} on each image of the dataset in use. To ensure the high quality of 3D reconstruction, we employed the LSFM \cite{both2016lsfm} identity model concatenated with the expression model of 4DFAB \cite{ChengCVPR18}. The 4DFAB expression model was built from a collection of over 10,000 expressive face 3D scans of spontaneous and posed expressions, collected from 180 individuals in 4 sessions over the period of 5 years. SliderGAN exploits the scale and representation power of 4DFAB to learn how to realistically edit facial expressions in images. The method described above constitutes a technique to automatically annotate the dataset and eliminates the need of costly manual annotation. 

\textbf{3D Warped Images} One crucial problem of training with pseudo-annotations extracted by 3DMM fitting on images, is that the parameter values are not always consistent as small variations in expression can be mistakenly explained by the identity, texture or camera model of the 3DMM. To overcome this limitation, we augment the training dataset with expressive images that we render and therefore know the exact blenshape parameter values. In more detail, we fit with the same 3DMM 10,000 images of EmotioNet in order to recover the identity and camera models for each image. A 3D texture can also be sampled by projecting the recovered mesh on the original image. Then, we combined the identity meshes with randomly generated expressions from the 4DFAB expression model and rendered back on the original images. Rendering 20 different expressions from each image, we augmented the dataset by 200,000 accurately annotated images. Some of the generated images are displayed in Fig. \ref{fig_rendered_data}

\textbf{4DFAB Images} A common problem of developing generative models of facial expression is the difficulty in accurately measuring the quality of the generated images. This is mainly due to the lack of databases with images of people of the same identity with arbitrary expressions. To overcome this issue and quantitatively measure the quality of images generated by SliderGAN, as well as compare with the baseline, we created a database with rendered images from 3D meshes and textures of 4DFAB. In more detail, we rendered 100 to 500 images with arbitrary expression from each of the 180 identities and for each of the 4 sessions of 4DFAB, thus rendering 300,000 images in total. To obtain expression parameters for each rendered image, we projected the blendshape model $\mathcal{S}_{exp}$ on each corresponding 3D mesh $\mathbf{S}$ such that the obtained parameters are $\mathbf{p} = \mathbf{U}^\top_{exp}(\mathbf{S} - \mathbf{\bar{s}})$.

\textbf{Lip Reading Words in 3D (LRW-3D)} Lip Reading in the Wild (LRW) dataset \cite{Chung16} consists of videos of hundreds of speakers including up to 1000 utterances of 500 different words. LRW-3D \cite{Tzirakis2019Synthesising3F} provides speech blendshapes parameters for the frames of LRW, which were recovered by mapping each frame of LRW that correspond to one of the 500 words to instances of a 3D blendhshape model of speech, by aligning the audio segments of the LRW videos and those of a 4D speech database. Moreover, to extract expression parameters for each word segment of the videos we applied the 3DMM video fitting algorithm of \cite{booth20183d}, which accounts for the temporal dependency between frames. In Section \ref{subsec_speech}, we utilize the annotations of LRW-3D as well as the expression parameters to perform expression and speech transfer.

\subsection{3D Model-based Expression Editing}\label{subsec_3dbased}


\textbf{Sliding single expression parameters} In this experiment we demonstrate the capability of SliderGAN to edit the facial expression of images when single expression parameters are slid within the normalized range [-1, 1]. In Fig. \ref{fig_single_param} we provide results for 10 levels of activation of single parameters of the model (-1, -0.8, -0.6, -0.4, -0.2, 0, 0.2, 0.4, 0.6, 0.8, 1), while the rest parameters remain zero. As can be observed in Fig. \ref{fig_single_param}, SliderGAN successfully learns to reproduce the behaviour of each blendshape separately, producing realistic facial expressions while maintaining the identity of the input image. Also, the transition between the generated expressions is smooth for successive values of the same parameter and the intensity of the expressions dependent on the magnitude of the parameter value. Note that when the zero vector is applied, SliderGAN produces the neutral expression, whatever the expression of the original image. 



\smallskip
\textbf{Sliding multiple expression parameters} The main feature of SliderGAN is its ability to edit facial expressions in images by sliding multiple parameters of the model, similarly to sliding parameters in a blendshape model to generate new expressions of a 3D face mesh. To test this characteristic of our model, we synthesize random expressions by conditioning the generator input on parameter vectors with elements randomly drawn from the standard normal distribution. Note that the model was trained with expression parameters normalized by the square root of the eigenvalues $e_i, i=1,...,N$ of the PCA blendshape model. This means that all combinations of expression parameters within the range [-1, 1] correspond to feasible facial expressions.

\begin{figure*}
    \hspace{9pt}
    \includegraphics[width=0.95\textwidth]{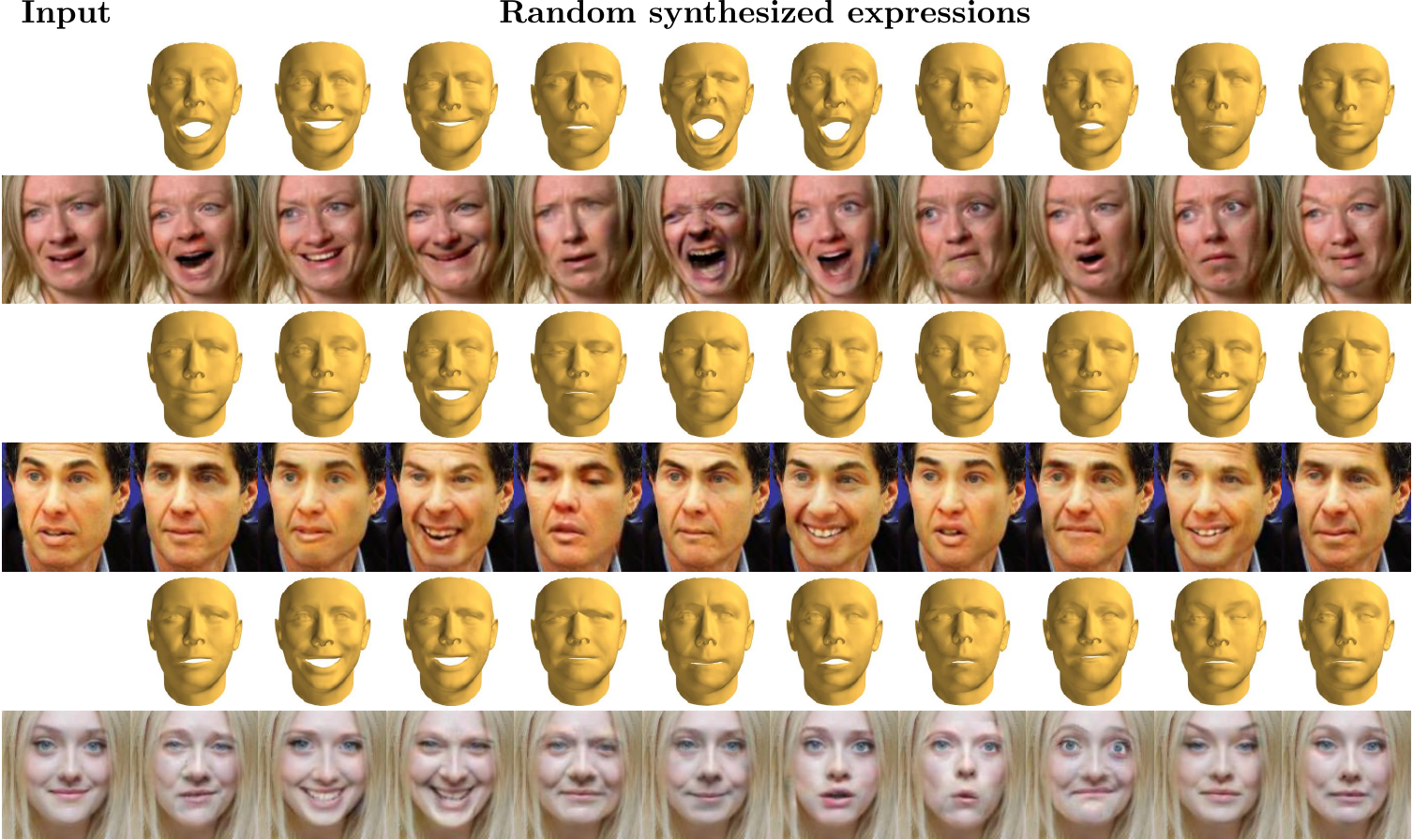}
    \caption{Expressive faces generated by sliding multiple blendshape (b/s) parameters in the range $[-1, 1]$. As it is observed, the wide range of the edited images accurately replicate the 3D faces' motion.}
    \label{fig_multi_param}
\end{figure*}

As illustrated by Fig. \ref{fig_multi_param}, SliderGAN is able to synthesize face images with a great variability of expressions, while maintaining identity. The generated expressions accurately resemble the 3D meshes' expressions when the same vector of parameters is used for the blendshape model. This fact makes our model ideal for facial expression editing in images. A target expression can first be chosen by utilizing the ease of perception of 3D visualization of a 3D blendshape model and then, the target parameters can be employed by the generator to edit a face image accordingly. 

\subsection{Expression Transfer and Interpolation}\label{subsec_exprtrsf}

A by-product of SliderGAN is that the discriminator $\mathcal{D}$ learns to map images to expression parameters $\mathcal{D}_\mathbf{p}$ that represent their 3D expression through $\mathcal{S}_{exp}(\mathcal{D}_\mathbf{p})$. We capitalize on this fact to perform direct expression transfer and interpolation between images without any annotations about expression. Assuming a source image $\mathbf{I}_{src}$ with expression parameters $\mathbf{p}_{src} = \mathcal{D}_\mathbf{p}(\mathbf{I}_{src})$ and a target image $\mathbf{I}_{trg}$ with expression parameters $\mathbf{p}_{trg} = \mathcal{D}_\mathbf{p}(\mathbf{I}_{trg})$, we are able to transfer expression $\mathbf{p}_{trg}$ to image $\mathbf{I}_{src}$ by utilising the generator of SliderGAN, such that $\mathbf{I}_{src \rightarrow trg} = \mathcal{G}(\mathbf{I}_{src}|\mathbf{p}_{trg})$. Note that no 3DMM fitting or manual annotation is required to extract the expression parameters and transfer the expression, as this is performed by the trained discriminator. 

Additionally, by interpolating the expression parameters of the source and target images, we are able to generate expressive faces that demonstrate a smooth transition from expression $\mathbf{p}_{src}$ to expression $\mathbf{p}_{trg}$. Interpolation of the expression parameters can be performed by sliding an interpolation factor $a$ within the region [0,1] such that the requested parameters are $\mathbf{p}_{interp} = a\mathbf{p}_{src} + (1 - a)\mathbf{p}_{trg}$.

\begin{figure*}
    \includegraphics[width=\textwidth]{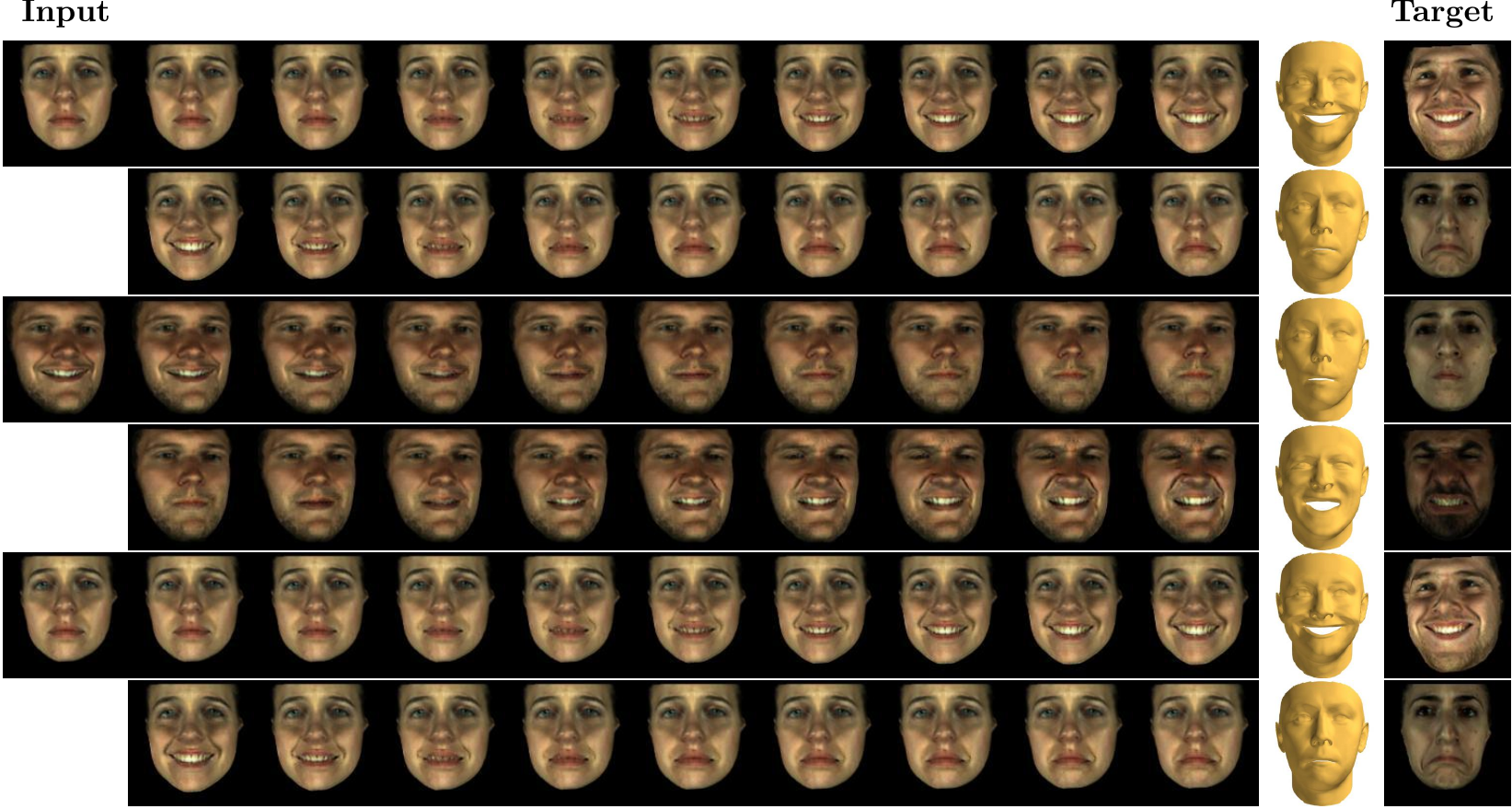}
    \caption{Expression interpolation between images of 4DFAB. First, we employ $\mathcal{D}$ to recover the expression parameters from an input and the target images. Then, we capitalize on these parameter vectors to animate the expression of the input image towards multiple targets.}
    \label{fig_exprtrsf_4dfab}
\end{figure*}
\begin{figure*}
    \includegraphics[width=\textwidth]{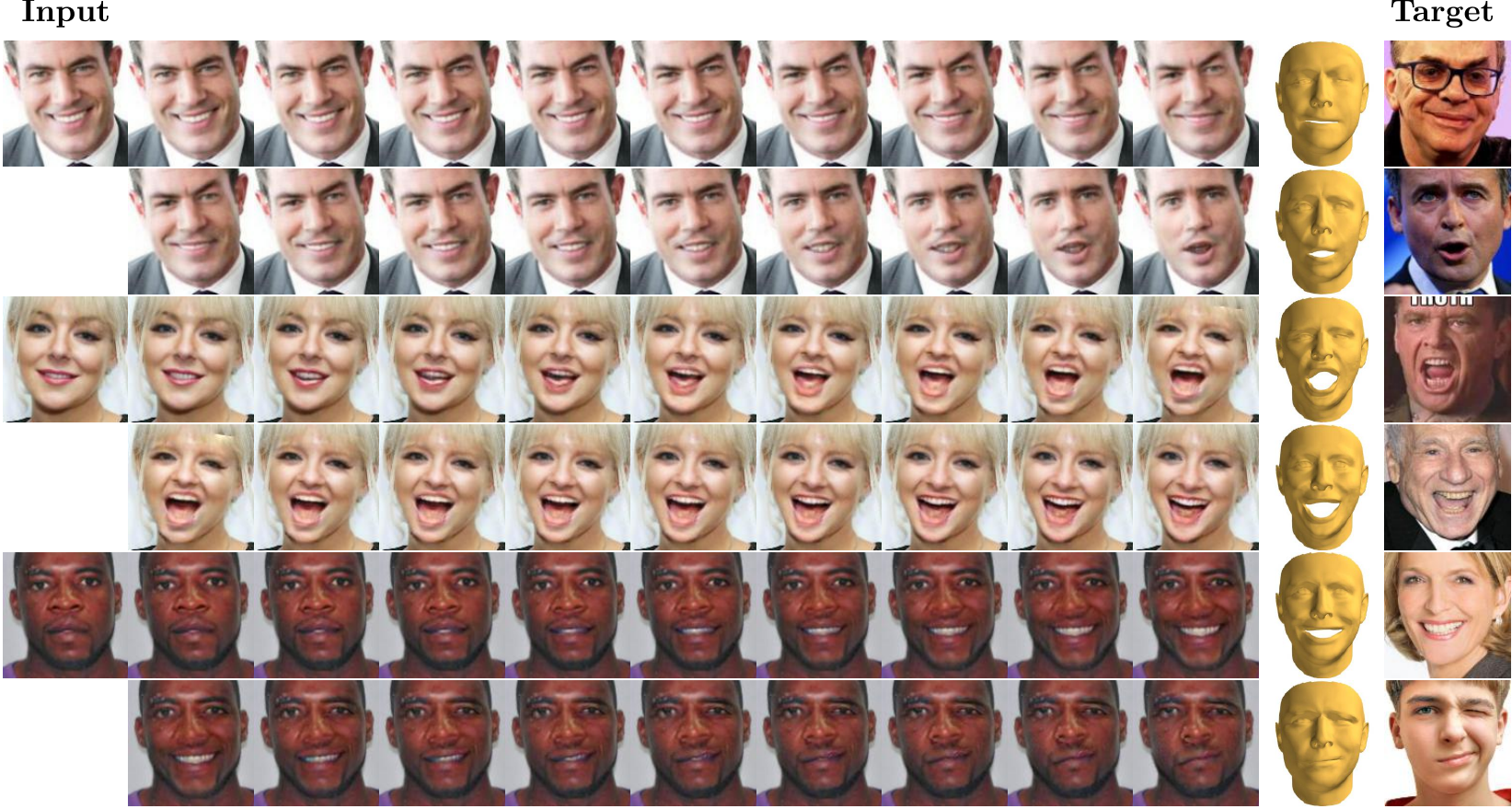}
    \caption{Expression interpolation between images of Emotionet. First, we employ $\mathcal{D}$ to recover the expression parameters from an input and the target images. Then, we capitalize on these parameter vectors to animate the expression of the input image towards multiple targets.}
    \label{fig_exprtrsf_emo}
\end{figure*}

\smallskip
\textbf{Qualitative Evaluation} Results of performing expression transfer and interpolation on images of the 4DFAB rendered database and Emotionet are displayed in Fig. \ref{fig_exprtrsf_4dfab} and Fig. \ref{fig_exprtrsf_emo} respectively, where it can be seen that the expressions of the generated images obviously reproduce the target expressions. The smooth transition between expressions $\mathbf{p}_{src}$ and $\mathbf{p}_{trg}$ indicates that SliderGAN successfully learns to map images to expressions across the whole expression parameter space. Also, it is evident that $\mathcal{D}$ accurately regresses the blendshape parameters from images $\mathbf{I}_{trg}$ by observing the recovered 3D faces. The accuracy of the regressed parameters is also examined in Section \ref{subsec_expr3Drec}.

\begin{figure*}
    \includegraphics[width=\textwidth]{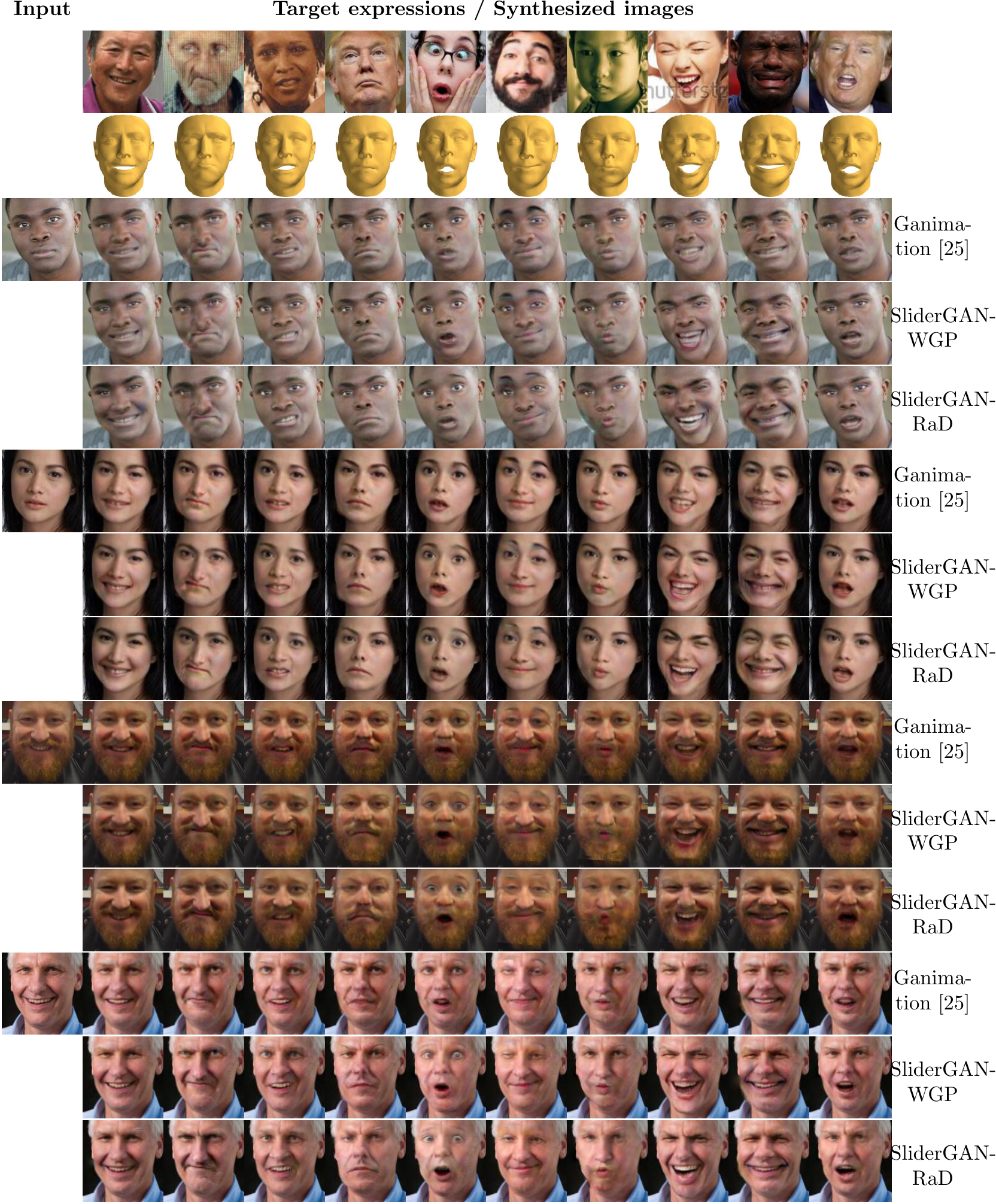}
    \caption{Expression transfer between images of Emotionet. First, we employ $\mathcal{D}$ to recover expression parameters from the target images. Then, we utilize these parameter vectors to transfer the target expressions to the input images. From the results, SliderGAN-RaD produces higher quality textures than any of the other two methods (mostly evident in the mouth and eyes regions). Moreover, GANimation reproduces the target expressions with lower accuracy. (Please, zoom in the images to notice the differences in texture quality.)}
    \label{fig_exprtrsf_comparison}
\end{figure*}

To further validate the quality of our results, we trained GANimation on the same dataset with AU annotations extracted with OpenFace \cite{amos2016openface} as suggesed by the authors. We performed expression transfer between images and present results for SliderGAN-RaD, SliderGAN-WGP and GANimation. In Fig. \ref{fig_exprtrsf_comparison}, it is obvious that SliderGAN-RaD benefits from the Relativistic GAN training and produces higher quality textures than SliderGAN-WGP, while both SliderGAN implementations better simulate the expressions of the target images than GANimation.

\smallskip
\textbf{Quantitative Evaluation} In this section we provide quantitative evaluation on the performance of SliderGAN on arbitrary expression transfer. We employ the 4DFAB rendered images dataset which allows us to calculate the Image Euclidean Distance \cite{Wang2005EDI} between ground truth rendered images of 4DFAB and images generated by SliderGAN. Image Euclidea Distance is a robust alternative metric to the standard pixel loss for image distances, which is defined between two RGB images $x$ and $y$ each with $M \times N$ pipxels as: \begin{equation}
\abovedisplayskip=\eqsep
\belowdisplayskip=\eqsep
    \frac{1}{2\pi} \sum_{i=1}^{MN}\sum_{j=1}^{MN} \exp\{|P_i - P_j|^2 / 2\}(\|x_i - y_i\|^2)(\|x_j - x_j\|^2)
\label{eq:img_euclidean_dst}
\end{equation}
where $P_i$ and $P_j$ are the pixel locations on the 2D image plane and $x_i, y_i, x_j, y_j$ the RGB values of images $x$ and $y$ at the vectorized locations $i$ and $j$. 

We trained SliderGAN with the rendered images from 150 identities of 4DFAB, leaving 30 identities for testing. To allow direct comparison between generated and real images, we randomly created 10,000 pairs of images of the same session and identity (this ensures that the images were rendered with the same camera conditions) from the testing set and performed expression transfer within each pair. To compare our model against the baseline model GANimation, we trained and performed the same experiment using GANimation on the same dataset with AUs activations that we obtained with OpenFace. Also, to showcase the benefits of the relativistic discriminator in image quality of the generated images, we repeated the experiment with SliderGAN-WGP. The results are presented in Table \ref{tab_IED_4dfab} where it can be seen that SliderGAN-RaD produces images with the lowest IED.

\begin{table}
    \renewcommand{\arraystretch}{1.3}
    \centering
    \caption{Image Euclidean Distance (IED), calculated between ground truth images of 4DFAB and corresponding generated images by Ganimation~\cite{pumarola2018ganimation}, SliderGAN-WGP and SliderGAN-RaD. Results from SliderGAN-RaD produce the lowest IED between the three methods.}
    \begin{tabular}{cc}
        \toprule[1.5pt]
        Method & \textbf{IED} \\
        \midrule
        GANimation~\cite{pumarola2018ganimation} & $1.04e-02$ \\
        SliderGAN-WGP & $7.932-03$ \\
        SliderGAN-RaD & $\mathbf{6.84e-03}$ \\
        \bottomrule[1.25pt]
    \end{tabular}
    \label{tab_IED_4dfab}
\end{table}

\begin{figure*}
    \hspace{80pt}
    \includegraphics[width=0.65\textwidth]{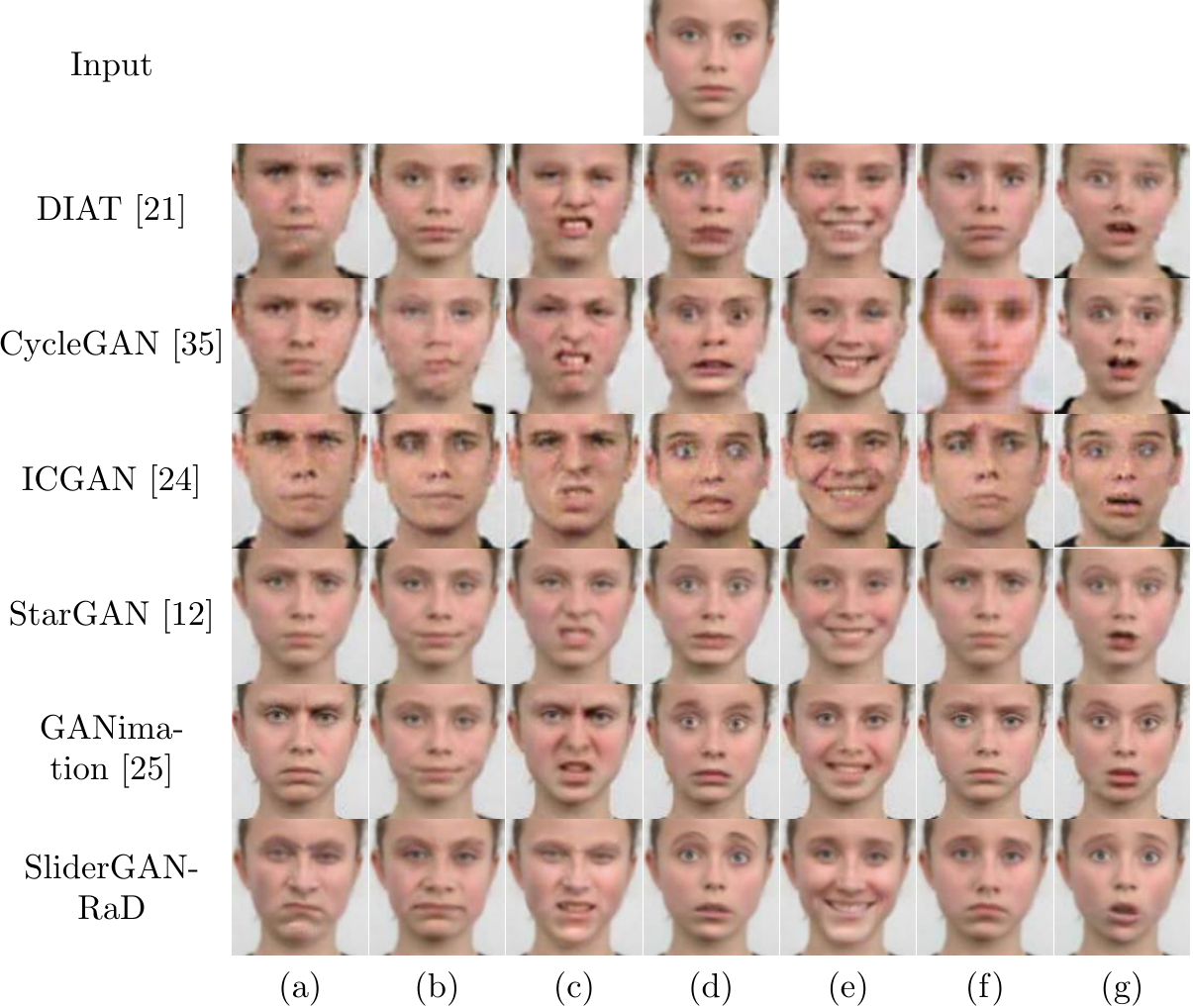}
    \caption{Generation of the 7 discrete expressions a) anger, b) contempt, c) disgust, d) fear, e) happiness, f) sadness, g) surprise. By comparing SliderGAN against DIAT \cite{LiZZ16e}, CycleGAN \cite{CycleGAN2017}, IcGAN \cite{IcGAN}, StarGAN \cite{StarGAN2018} and GANimation \cite{pumarola2018ganimation} we observe that our model generates results of high texture quality that resemble the queried expressions. The results of the rest of the methods where taken from \cite{pumarola2018ganimation}.}
    \label{fig_discrete}
\end{figure*}

\begin{table*}
    \renewcommand{\arraystretch}{1.3}
    \centering
    \caption{Expression recognition results by applying the off-the-self expression recognition system \cite{li2017reliable} of images generated by GANimation~\cite{pumarola2018ganimation}, SliderGAN-WGP and SliderGAN-RaD. Accuracy scores from both SliderGAN models outperform those of GANimation, while SliderGAN-RaD achieves thehighest accuracy in all epressions. }
    \begin{tabular}{ccccccccc}
        \toprule[1.5pt]
        Method & \textbf{Anger} & \textbf{Disgust} & \textbf{Fear} & \textbf{Happiness} & \textbf{Sadness} & \textbf{Surprise} & \textbf{Neutral} & \textbf{Average} \\
        \midrule
        GANimation~\cite{pumarola2018ganimation} & 0.552 & 0.446 & 0.517 & 0.658 & 0.632 & 0.622 & 0.631 & 0.579 \\
        SliderGAN-WGP & 0.550 & 0.463 & 0.514 & 0.762 & 0.633 & 0.678 & 0.702 & 0.614 \\
        SliderGAN-RaD & \textbf{0.591} & \textbf{0.481} & \textbf{0.531} & \textbf{0.798} & \textbf{0.654} & \textbf{0.689} & \textbf{0.708} & \textbf{0.636} \\
        \bottomrule[1.25pt]
    \end{tabular}
    \label{tab_expr_rec}
\end{table*}

\begin{figure}
    \includegraphics[width=0.48\textwidth]{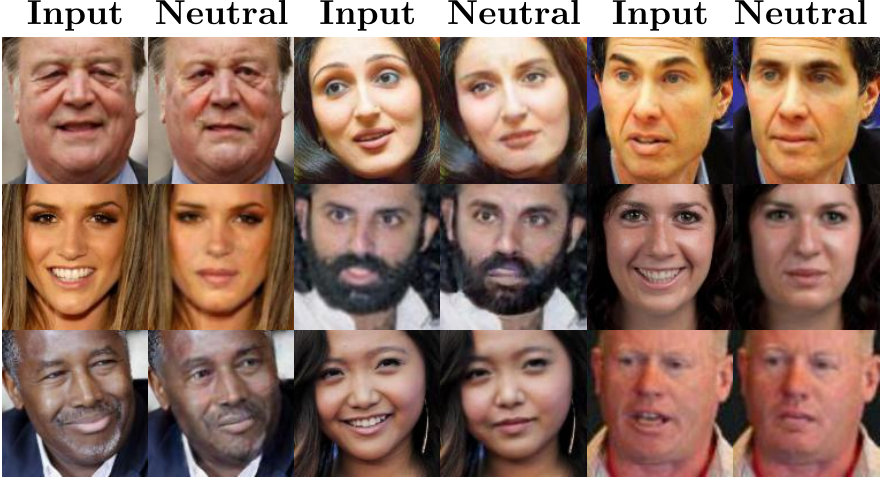}
    \caption{Neutralization of in-the-wild images of arbitrary expression. The neutralization takes place by setting all blendshape parameter values to zero.}
    \label{fig_neutral}
\end{figure}

\begin{figure*}
    \includegraphics[width=\textwidth]{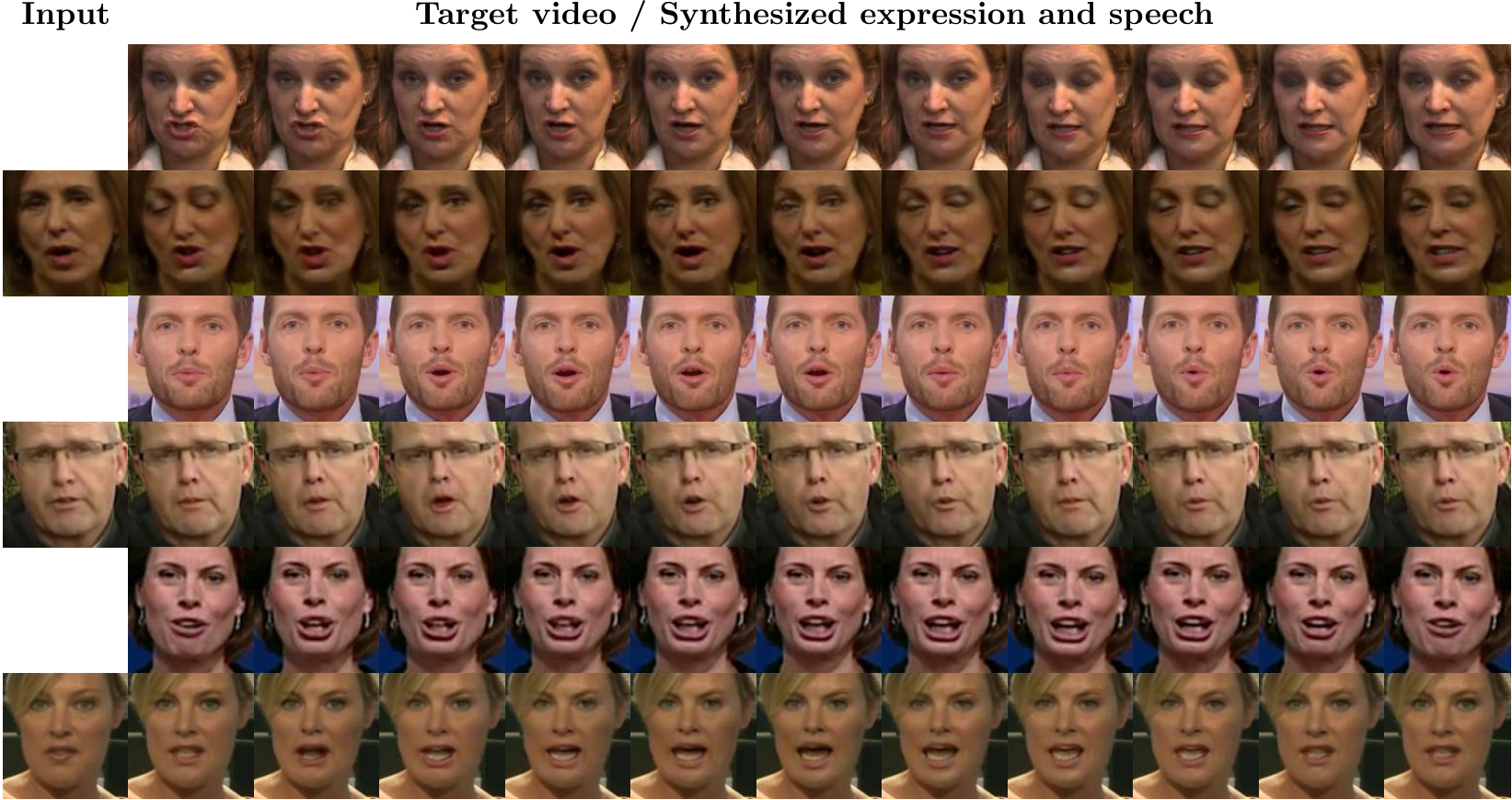}
    \caption{Combined expression and speech animation from a single input image. We utilize as targets the expression and speech blendshape parameters of consecutive frames of videos of LRW, to synthesize sequences of expression and speech from a single input image.}
    \label{fig_speech_trsf}
\end{figure*}

\subsection{Synthesis of Discrete Expressions}\label{subsec_six_basic} 

Specific combinations of the 3D expression model parameters represent the discrete expressions anger, contempt, fear, disgust, happiness, sadness, surprise and neutral. We employ these parameter vectors to synthesize expressive face images of the aforementioned discrete expressions and test our results both qualitatively and quantitatively.


\smallskip
\textbf{Qualitative Evaluation} To evaluate the performance of SliderGAN in this task, we visually compare our results against the results of five baseline models: DIAT \cite{LiZZ16e}, CycleGAN \cite{CycleGAN2017}, IcGAN \cite{IcGAN}, StarGAN \cite{StarGAN2018} and GANimation \cite{pumarola2018ganimation}. In Fig. \ref{fig_discrete} it is evident that SliderGAN generates results that resemble the queried expressions while maintaining the original face's identity and resolution. The results are close to those of GANimation, however the Relativistic GAN training of SliderGAN allows for slightly higher quality of images.

The neutral expression can also be synthesized by SliderGAN when all the elements of the target parameter vector are set to $0$. In fact, the neutral expression of the 3D blendshape model is also synthesized by the same vector. Results of image neutralization on in-the-wild images of arbitrary expression are presented in Fig. \ref{fig_neutral}, where it can be observed that the neutral expression is generated without significant loss in faces' identity.


\smallskip
\textbf{Quantitative Evaluation} We further evaluate the quality of the generated expressions by performing expression recognition with the off-the-self recognition system \cite{li2017reliable}. In more detail, we randomly selected 10,000 images from the test set of Emotionet, translated them to each of the discrete expressions anger, disgust, fear, happiness, sadness, surprise, neutral and passed them to the expression recognition network. For comparison, we repeated the same experiment with SliderGAN-WGP and GANimation using the same image set. In Table \ref{tab_expr_rec} we report accuracy scores for each expression class separately, as well as the average accuracy score for the three methods. The classification results are similar for the three models, with both implementations of SliderGAN producing slightly higher scores, which demotes that GANimation's results include more fail cases.

\begin{figure*}
    \includegraphics[width=\textwidth]{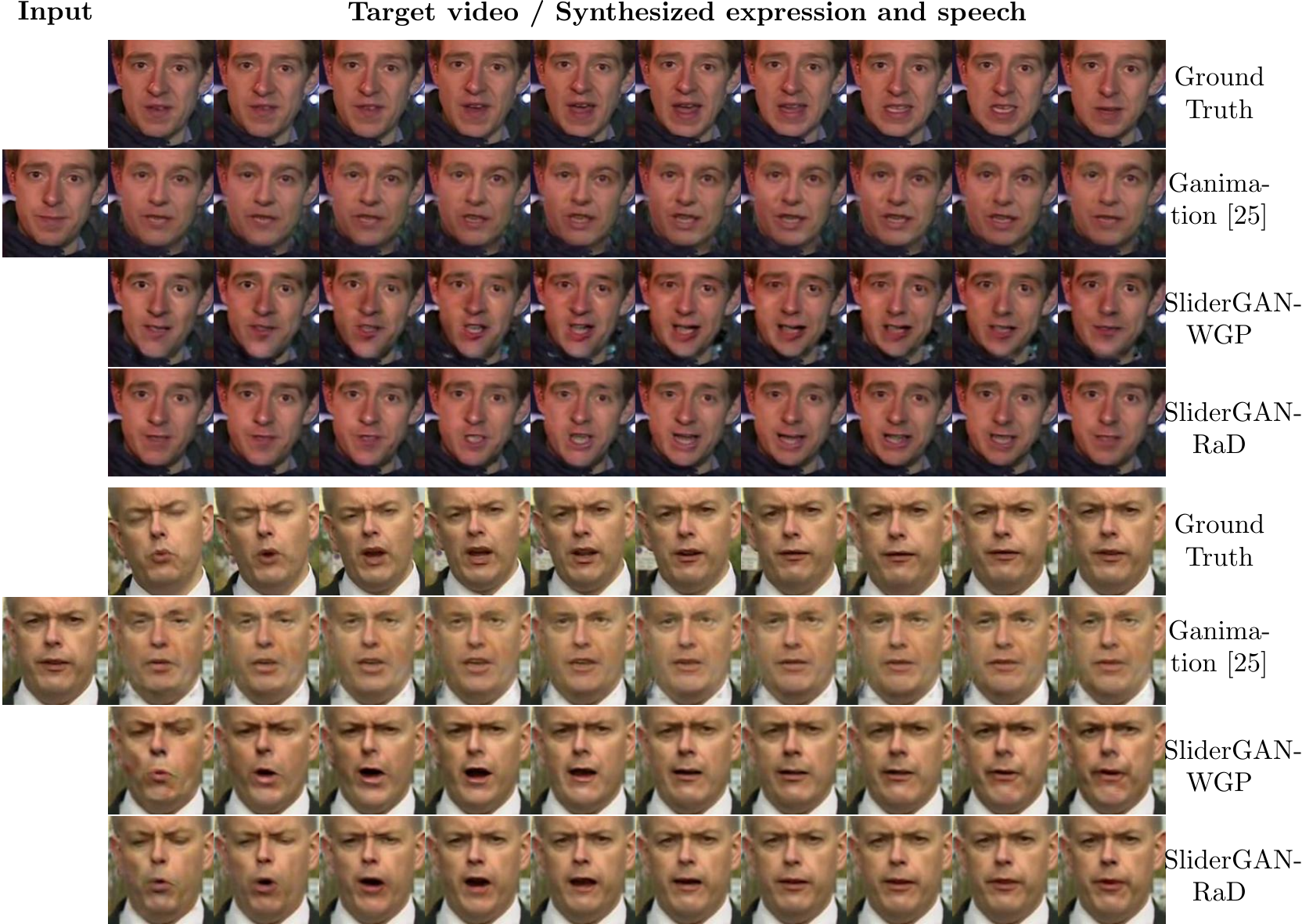}
    \caption{Comparison of combined expression and speech animation from a single input image between GANimation~\cite{pumarola2018ganimation}, SliderGAN-WGP and SliderGAN-RaD. We utilize as targets the expression and speech blendshape parameters of consecutive frames of a video of LRW. Then we reconstruct the expression and speech from a single input frame of the same video. Both SliderGAN implementations reconstruct face motion more accurately than GANimation. Also, the texture quality of the results is higher in SLiderGAN-RaD than in SLiderGAN-WGP as expected. (Please, zoom in the images to notice the differences in texture quality.)}
    \label{fig_speech_rec}
\end{figure*}

\subsection{Combined Expression and Speech Synthesis and Transfer}\label{subsec_speech}

Blendshape coding of facial deformations allows modelling arbitrary deformations (e.g. deformations due to identity, speech, non-human face morphing etc.) that are not limited to facial expressions, unlike AUs coding which is a system that taxonomizes the human facial muscles \cite{ekman2002facial}. Even though AUs 10-28 model mouth and lip motion, not all the details of lip motion that takes place during speech can be captured by these AUs. Moreover, only 10 (10, 12, 14, 15, 17, 20, 23, 25, 26, 28) out of these 18 AUs can automatically be recognized, which is achieved only with low accuracy. On the contrary, a blendshape model of the 3D motion of the human mouth and lips would better capture motion during speech, while it would allow the recovery of robust representations from images and videos of human speech.

We capitalize on this fact and employ the mouth and lips blendshape model of \cite{Tzirakis2019Synthesising3F} to perform speech synthesis from a single image with SliderGAN. Particularly, we employ the LRW-3D database which contains speech blendshape parameters annotations for the 500 words of LRW \cite{Chung16}, to perform combined expression and speech synthesis and transfer, which we evaluate both qualitatively and quantitatively.


\smallskip
\textbf{Qualitative Evaluation} LRW contains videos with both expression and speech. Thus, to completely capture the smooth face motion across frames we employed 30 expression parameters recovered by 3DMM fitting and 10 speech parameters of LRW-3D which correspond to the ten most significant components of the 3D speech model. We trained SliderGAN with 180,000 frames of LRW, without leveraging the temporal characteristics of the database, that is we shuffled the frames and trained our model with random target vectors to avoid learning person specific deformations. Results of performing expression and speech synthesis from a video using a single image are presented in Fig. \ref{fig_speech_trsf} where the the parameters and the input frame belong to the same video (ground truth frames are available) and in Fig. \ref{fig_speech_rec} where the parameters and the input frame belong to different videos of LRW.

For comparison we trained GANimation on the same dataset with AU activations obtained by OpenFace. As can be seen by Fig. \ref{fig_speech_trsf} and Fig. \ref{fig_speech_rec}, GANimation is not able to accurately simulate the lip motion of the target video. On the contrary, SliderGAN-WGP simulates mouth and lip motion well, but produces textures that look less realistic. SliderGAN-RaD produces higher quality results that look realistic in terms of accurate deformation and texture.

\smallskip
\textbf{Quantitative Evaluation} To measure the performance of our model we employ Image Euclidean Distance (IED) \cite{Wang2005EDI} to evaluate the results of expression and speech synthesis when the input frame and target parameters belong to the same video sequence. Due to changes in pose in the target videos, we align all target frames with the corresponding output ones before calculating IED. The results are presented in Table \ref{tab_IED_bbc}, where it can be seen that SliderGAN-RaD achieves the lowest error.

\begin{table}
    \renewcommand{\arraystretch}{1.3}
    \centering
    \caption{Image Euclidean Distance (IED), calculated between ground truth images of LRW and corresponding generated images by Ganimation~\cite{pumarola2018ganimation}, SliderGAN-WGP and SliderGAN-RaD. Results from SliderGAN-RaD produce the lowest IED between the three methods, which indicates the robustness of blendshape coding for speech utlized by SliderGAN.}
    \begin{tabular}{cc}
        \toprule[1.5pt]
        Method & \textbf{IED} \\
        \midrule
        GANimation~\cite{pumarola2018ganimation} & $3.07e-02$ \\
        SliderGAN-WGP & $1.14e-02$ \\
        SliderGAN-RaD & $\mathbf{9.35e-03}$ \\
        \bottomrule[1.25pt]
    \end{tabular}
    \label{tab_IED_bbc}
\end{table}

\subsection{3D Expression Reconstruction}\label{subsec_expr3Drec}

As also described in Section \ref{subsec_exprtrsf}, a by-product of SliderGAN is the discriminator's ability to map images to expression parameters $\mathcal{D}_\mathbf{p}$ that reconstruct the 3D expression as $\mathcal{S}_{exp}(\mathcal{D}_\mathbf{p})$. We test the accuracy of the regressed parameters on images of Emotionet in two scenarios: a) we calculate the error between parameters recovered by 3DMM fitting and those regressed by $\mathcal{D}$ on the same image as (Table \ref{tab_error_expr_params} row 1) and b) we test the consistency of our model and calculate the error between some target parameters $\mathbf{p}_{trg}$ and those regressed by $\mathcal{D}$ on a manipulated image which was translated to expression $\mathbf{p}_{trg}$ by SliderGAN-RaD (Table \ref{tab_error_expr_params} row 2).

For comparison, we repeated the same experiment with GANimation for which we calculated the errors in AUs activations. For both experiments we employed 10000 images from our test set. The results demonstrate that the discriminator of SliderGAN-RaD extracts expression parameters from images with high accuracy compared to 3DMM fitting. On the contrary, GANimation's discriminator is less consistent in recovering AU annotations when compared to those of OpenFace. This, also, illustrates that the robustness of blendshape coding of expression over AUs, makes SliderGAN more suitable than GANimation for direct expression transfer. 

\begin{table}
    \renewcommand{\arraystretch}{1.3}
    \centering
    \caption{Expression representation results on SLiderGAN-RaD (blendshape parameters coding) and Ganimation (AUs activations coding). SliderGAN is capable to accurately and robustly recover expression representations, while GANimation fails to detect AUs activations.}
    \begin{tabular}{ccc}
        \toprule[1.5pt]
         & SliderGAN & GANimation~\cite{pumarola2018ganimation}
        \\
        \midrule
        \vspace{-10pt}
        \\
        \smallskip\smallskip
        \hspace{-10pt} $\frac{1}{N}\sum_{i=1}^{N}\frac{{\|\mathbf{p}_{3DMM,i} - \mathbf{p}_{D,i}\|}}{\|\mathbf{p}_{3DMM,i}\|}$ & \textbf{0.131} & 0.427
        \\
        \smallskip
        \hspace{-10pt} $\frac{1}{N}\sum_{i=1}^{N}\frac{{\|\mathbf{p}_{trg,i} - \mathbf{p}_{D,i}\|}}{\|\mathbf{p}_{trg,i}\|}$ & \textbf{0.258} & 0.513
        \\
        \bottomrule[1.25pt]
    \end{tabular}
    \label{tab_error_expr_params}
\end{table}

\begin{figure}
    \includegraphics[width=0.48\textwidth]{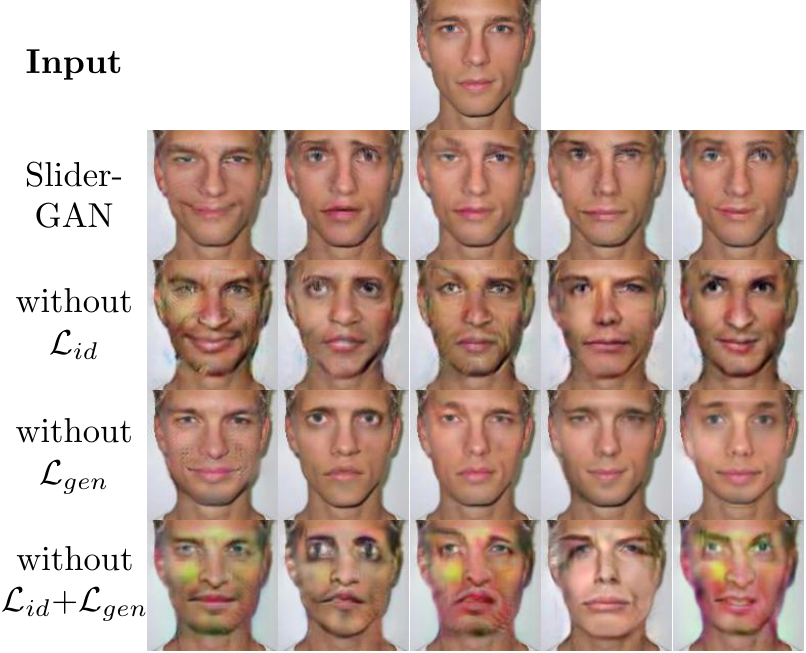}
    \caption{Results from the ablation study on SliderGAN's loss function components. It is evident that both losses $\mathcal{L}_{id}$ and $\mathcal{L}_{gen}$ have significant impact on the training of the model, with $\mathcal{L}_{id}$ being the most important for generating realistic images.}
    \label{fig_ablation}
\end{figure}

\subsection{Ablation Study}\label{subsec_ablation}

In this section we investigate the effect of the different losses that constitute the total loss functions $\mathcal{L}_{\mathcal{G}}$ and $\mathcal{L}_{\mathcal{D}}$ of our algorithm. As discussed in Section \ref{subsec_model}, both training in a semi-supervised manner with loss $\mathcal{L}_{gen}$ and employing a face recognition loss $\mathcal{L}_{id}$ between the original and the generated images, contribute significantly in the training process of the generator $\mathcal{G}$. To explore the extend at which these losses improve or affect the performance of $\mathcal{G}$, we consider three different models trained with variations of the loss function of SliderGAN which are: a) $\mathcal{L}_{\mathcal{G}}$ does not include $\mathcal{L}_{id}$, b) $\mathcal{L}_{\mathcal{G}}$ does not include $\mathcal{L}_{gen}$ and c) $\mathcal{L}_{\mathcal{G}}$ does not include both $\mathcal{L}_{id}$ and $\mathcal{L}_{id}$. Fig. \ref{fig_ablation} depicts results for the same subject generated by the three models as well as SliderGAN. As it can be observed, the absence of $\mathcal{L}_{id}$ affects the quality of the generated images more, as more artifacts are produced. However, $\mathcal{L}_{gen}$ vitally supports $\mathcal{L}_{id}$ in accurately simulating the target expression and producing good quality textures. When both $\mathcal{L}_{id}$ and $\mathcal{L}_{id}$ are omitted, both the identity preservation and the expression accuracy decrease drastically.

\section{Conclusion}\label{section_conclusion}

In this paper, we presented SliderGAN, a new and very flexible way for manipulating the expression (i.e., expression transfer etc.) in facial images driven by a set of statistical blendshapes. To this end, a novel generator based on Deep Convolutional Neural Networks (DCNNs) is proposed, as well as a learning strategy that makes use of adversarial learning. A by-product of the learning process is a very powerful regression network that maps the image into a number of blenshape parameters, which can then be used for conditioning the inputs of the generator.

{
\bibliographystyle{spmpsci}      
\bibliography{egbib}
}

\end{document}